\documentclass[journal]{IEEEtran}

\ifCLASSINFOpdf
\else
\fi
\usepackage{graphicx}
\usepackage{booktabs}
\usepackage{amsmath}
\usepackage{rotating}
\usepackage[pagebackref=true,breaklinks=true,colorlinks,linkcolor=blue,citecolor=blue]{hyperref}
\usepackage{color}
\usepackage{times}
\usepackage{epsfig}
\usepackage{graphicx}
\usepackage{amssymb}
\usepackage{multirow}
\usepackage{algorithm}
\usepackage{algorithmic}
\usepackage{array}
\usepackage{cite}
\newcommand{\RN}[1]{%
  \textup{\uppercase\expandafter{\romannumeral#1}}%
}

\usepackage[utf8]{inputenc}
\usepackage[T1]{fontenc}
\usepackage{textcomp}
\usepackage[table]{xcolor}
\usepackage{threeparttable}

\begin{document}

\title{Cross-Skeleton Interaction Graph Aggregation Network for Representation Learning of Mouse Social Behaviour}
\author{Feixiang Zhou, Xinyu Yang, Fang Chen, Long Chen, Zheheng Jiang, Hui Zhu, Reiko Heckel, Haikuan Wang, Minrui Fei and Huiyu Zhou % <-this % stops a space
\thanks{F. Zhou is with School of Eye \& Vision Sciences, University of Liverpool, United Kingdom.}
\thanks{X. Yang, F. Chen, Z. Jiang, R. Heckel and H. Zhou are with School of Computing and Mathematical Sciences, University of Leicester, United Kingdom. H. Zhou is the corresponding author. E-mail: hz143@leicester.ac.uk. }
% <-this % stops a space
\thanks{ L. Chen is with Institute of Clinical Sciences, Faculty of Medicine, Imperial College London, United Kingdom.}
\thanks{H. Zhu is with School of Electronics and Information, Jiangsu University of Science and Technology, China}
\thanks{H. Wang and M. Fei are with School of Mechatronic Engineering and Automation, Shanghai University, China}
}
% \markboth{IEEE Transactions on image processing}
% {Shell \MakeLowercase{\textit{et al.}}: Bare Demo of IEEEtran.cls for IEEE Journals}

\maketitle

\begin{abstract}
Automated social behaviour analysis of mice has become an increasingly popular research area in behavioural neuroscience. Recently, pose information (i.e., locations of keypoints or skeleton) has been used to interpret social behaviours of mice. Nevertheless, effective encoding and decoding of social interaction information underlying the keypoints of mice has been rarely investigated in the existing methods. In particular, it is challenging to model complex social interactions between mice due to highly deformable body shapes and ambiguous movement patterns. To deal with the interaction modelling problem, we here propose a Cross-Skeleton Interaction Graph Aggregation Network (CS-IGANet) to learn abundant dynamics of freely interacting mice, where a Cross-Skeleton Node-level Interaction module (CS-NLI) is used to model multi-level interactions (i.e., intra-, inter- and cross-skeleton interactions). Furthermore, we design a novel Interaction-Aware Transformer (IAT) to dynamically learn the graph-level representation of social behaviours and update the node-level representation, guided by our proposed interaction-aware self-attention mechanism. Finally, to enhance the representation ability of our model, an auxiliary self-supervised learning task is proposed for measuring the similarity between cross-skeleton nodes. Experimental results on the standard CRMI13-Skeleton and our PDMB-Skeleton datasets show that our proposed model outperforms several other state-of-the-art approaches.
\end{abstract}

\begin{IEEEkeywords}
Social behaviour recognition, Graph neural network, Self-attention, Self-supervision,  Cross-skeleton.
\end{IEEEkeywords}

% encourage the proposed model to focus on the similarity and dissimilarity between cross-skeleton node pairs


\IEEEpeerreviewmaketitle

\section{Introduction}
\label{section1}
\IEEEPARstart{T}{H}{E} analysis of rodent social behaviour is an interesting issue in neuroscience and pharmacology. Laboratory mice provide a valuable platform to study psychiatric and neurological disorders such as Huntington’s \cite{urbach2014automated}, Alzheimer’s \cite{Lewejohann2009}, schizophrenia \cite{wilson2014social}, as well as Parkinson’s disease (PD) \cite{Blume2009} because mice offer several advantages, including their genetic similarity to humans and the ability to manipulate and control experimental conditions. Traditionally, social behaviour identification is performed by manually annotating hours of video recordings of interactions between mice with pre-defined behaviour labels. Unfortunately, this human labelling practice is time-consuming, error-prone and highly subjective. Recent advances in computer vision and pattern recognition have facilitated automated analysis of mouse behaviours \cite{jiang2018context,jiang2021muti, arac2019deepbehavior,Nguyen2019,marks2022deep,camilleri2023mice}, which provides another dimension to understand the relationship between neural activities and behavioural phenotypes in neuroscience research.

% One type of previous systems works on sensor-based approaches that involves adopting various equipment to monitor animal behaviours, including the use of radio-frequency identification (RFID) \cite{peleh2019rfid}, piezoelectric sensors \cite{flores2007pattern} and passive infrared (PIR) motion sensors \cite{brown2016compass}. However, these methods may have the invasiveness of the tag implantation, and limited capability to handle more complex mouse behaviours such as eating, attacking, or sniffing \cite{jiang2021muti}. Another alternative is video-based methodologies, where depth camera \cite{hong2015automated}, single- or multi-camera systems \cite{jiang2018context, DeChaumont2012, burgos2012social, Robie2017} have been successfully developed for qualifying mouse behaviours. Most of these approaches rely on visual features extracted from continuous RGB or gray images to train classifiers, which have yielded promising results.

Mouse social behaviour recognition is non-trivial due to the intricate nature of not just individual behaviours but also the interactions of mice. Compared to human behaviours, mouse social interactions exhibit ambiguous movement patterns as these interactions can involve subtle cues and rapid movements, making it difficult to recognise and interpret specific behaviours. This motivates us to design a novel computer vision solution to analyse intricate movement patterns and social interactions of mice, which is valuable in fields like biobehavioral research \cite{moulin2021rodent,marks2022deep}.  Additionally, our study has potential implications in studying neurological diseases, e.g., PD \cite{jiang2021muti,Blume2009,decourt2021neuropsychiatric}. By investigating how PD affects social behaviours of mice, we can gain insights into its neurological underpinnings, potentially contributing to the early diagnosis of human diseases and the development of treatments or interventions. We believe that our research has the potential to yield valuable insights in the domains of image processing, neuroscience and biobehavioral research.

% The motivation behind choosing mice experiments for our study lies in the significance of understanding social behaviours of mice and their relevance to neurological disorders such as Parkinson's disease. Specifically, we aim to analyse intricate movement patterns and social interactions of mice from skeleton information by designing a novel computer vision solution, which is valuable in fields like biobehavioral research \cite{moulin2021rodent,marks2022deep}. Additionally, this research has potential implications in studying diseases like Parkinson's disease \cite{jiang2021muti,Blume2009,decourt2021neuropsychiatric}. By investigating how Parkinson's disease affects social behaviours of mice, we can gain insights into its neurological underpinnings, potentially contributing to the development of treatments or interventions. We believe that our research has the potential to yield valuable insights in the domains of image processing, neuroscience and biobehavioral research.
% Our work on analysing mouse social behaviour aims to provide a deeper understanding of social interactions between mice, which is valuable in fields like neuroscience and behavioural biology \cite{Blume2009}. 
% Additionally, it has potential implications in addressing diseases like Parkinson's disease. By comprehending how Parkinson's disease affects social behaviours in mice, we can gain insights into its neurological underpinnings, potentially contributing to the development of treatments or interventions.

As more and more accurate results have been provided by deep learning based pose estimation models \cite{Mathis2018, Graving2019,9492104,cao2019openpose}, researchers have started to directly recognise mouse social behaviours using pose information (i.e., the locations of keypoints generated by pose estimators) \cite{arac2019deepbehavior, nilsson2020simple}. Compared with RGB features, pose information includes only 2D or 3D positions of keypoints, which may be free of environmental noise (e.g. complex background and illumination changes) \cite{Liu2020}. However, the features extracted by most of the existing systems are hand-crafted, based on pre-defined keypoints. For instance, in \cite{nilsson2020simple},  distance relations between two noses (i.e., distance feature) are represented by the distance between the noses of two mice. Actually, such hand-crafted shallow features are insufficient to describe the dependency between the corresponding keypoints.
% since the spatial configuration of mouse body, i.e., skeleton, is not well explored. Actually, mouse skeleton, similar to the human skeleton (\citealt{yan2018spatial}), is naturally established as a graph structure in a non-Euclidean space, where the keypoints and their natural connections in the mouse body can be regarded as vertexes and edges respectively.
To this end, we need to develop an effective way to automatically model the spatio-temporal interactions between keypoints.

% The latest system called SimBa \citealt{nilsson2020simple} analyzes mouse social behaviours based on the original pose estimation tracking data where a random forests algorithm is leveraged to classify behavioural patterns. Although the SimBa provides an easy-to-use graphical user-interface (GUI) for users to directly analysis mouse social behaviours, the manually calculated shallow representations (e.g. area of mouse convex hull, distance between part1 and part2) limits it’s ability to interpret more complex social behaviours due to highly deformable body structures and diverse movement patterns. In particular, such  hand-crafted representations cannot fully express the spatio-temporal dependency between correlated joints since the mouse body structure, i.e., skeleton, is ignored. Actually, mouse skeleton, similar to the human skeleton \citealt{yan2018spatial}, is naturally established as a graph structure in a non-Euclidean space, where the joints and their natural connections in the mouse body can be regarded as vertexes and edges respectively. Similarly, in DeepBehaviour \citealt{arac2019deepbehavior}, several hand-crafted pose-based features (i.e., movement trajectories,  distance between the body centers of mice and velocities) are fed into neural networks for the analysis of social interactions of two mice, thus resulting in relatively weak behavioural representations. In addition, these systems fail to explore the correlation between the neighbouring behaviours in a long video recording.

Graph convolutional networks (GCNs) \cite{kipf2016semi}, which generalise convolution from images to graphs, have been successfully adopted in many areas to model graph-structured data, especially in skeleton based human action recognition approaches \cite{gupta2021quo, song2020stronger, xia2021multi,chi2022infogcn,hedegaard2023continual}. Nevertheless, most of GCN-based methods have been designed for action recognition of single objects rather than multiple interacting subjects. In most of these established models, a standard human skeleton with all joints is utilised to model the potential spatio-temporal dependencies between the joints. To capture discriminative action features, multi-view solutions \cite{wang2020learning} consisting of two ensemble models with different skeleton typologies are developed to utilise comprehensive information. Although such approach can significantly improve the discriminative capability, the two sub-models need to be trained independently - how to select a new and effective skeleton topology is difficult to determine. 
% t is imperative to consider the multi-scale and hierarchical nature of interactions
Moreover, to obtain the graph-level representation\footnote{In this paper, node-level representation refers to the features of each node provided in a graph. Graph-level representation refers to the overall features of the whole graph. } that represents a specific action, global average pooling (GAP) \cite{ shi2019two} is normally used to aggregate node-level representation from the final stage of the network. However, this operation processes all node features equally without considering the importance of different nodes, structural constraints and dependencies between them. This limitation results in a constrained ability to globally represent complex social interactions due to the ambiguous motion patterns of mice. Recently, GCN \cite{zhu2021dyadic,li2022two} and Transformer \cite{pang2022igformer,wen2023interactive} have been introduced for interactive action recognition. Although they consider intra-body and inter-body relations, the overall representation ability of the spatio-temporal interactions is still limited because of the above issue. Hence, how to effectively model complex and diverse social interactions of mice remains an open problem.

% multi-scale and hierarchical nature of interactions
% Moreover, to obtain the graph-level
%%For mouse social behaviour recognition, such graph-level representation inevitably results in the loss of significant interaction information due to highly deformable body structures and diverse movement patterns.

\begin{figure*}
\begin{center}
\includegraphics[width=17cm]{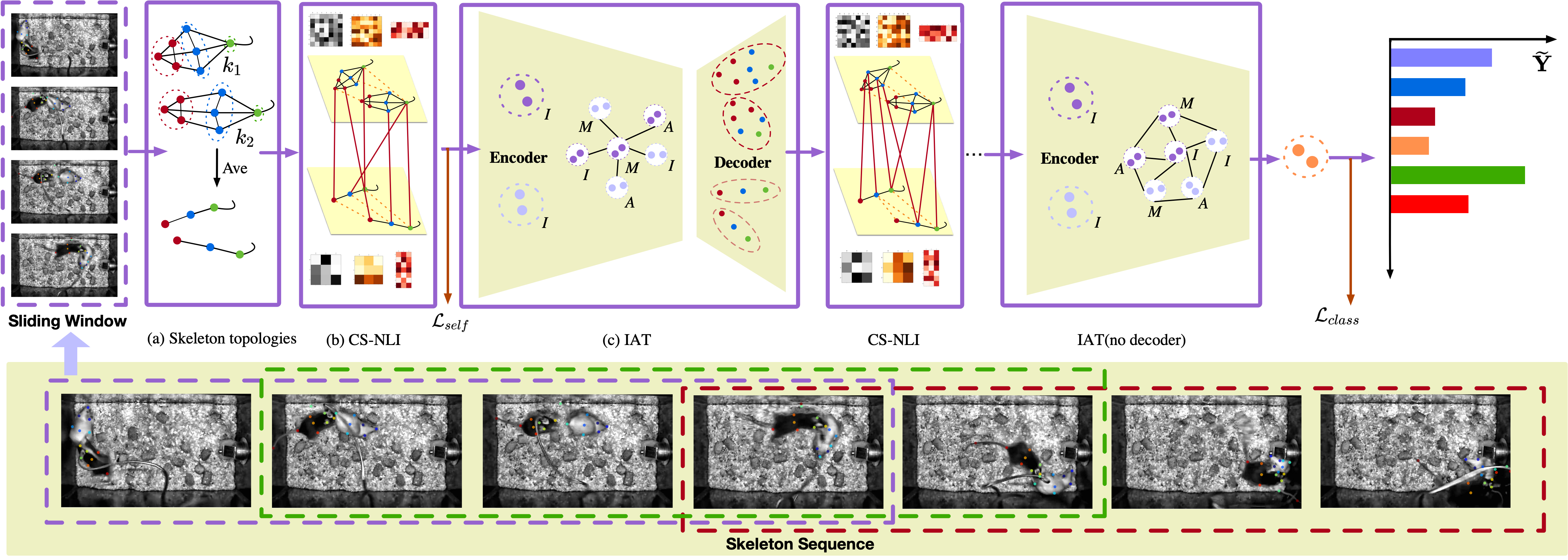}
\end{center}
\caption{Overview of the proposed CS-IGANet for mouse social behaviour representation learning. (a) The skeleton topologies of two mice. The input is a skeleton sequence from a long video, where each sliding window centered at a certain frame contains a specific behaviour \cite{jiang2021muti}. We define dense and sparse skeletons for describing diverse spatial structures of a mouse (Ave - average). (b) Cross-Skeleton Node-level Interaction (CS-NLI) module. It is composed of intra-skeleton (black solid lines and interaction matrix), inter-skeleton (yellow dashed lines) and cross-skeleton (red solid lines) interactions with the ability to reveal powerful node-level representation of social interactions (see Section \ref{section3.2}).
(c) Interaction-aware Transformer (IAT). For each skeleton branch, the node-level representation encoded by (b) is hierarchically aggregated to generate graph-level representation $I$ (The two dark purple circles represent the graph-level representations of two mice from the dense skeleton). In the encoder, we further fuse different representations ($I$, $M$ and $A$) to enhance graph-level representation (see Section \ref{section3.3.1}). The decoder aims to adaptively update the node-level representation. (see Section \ref{section3.3.2}). $\mathcal L_{class} $ is the cross-entropy loss, and $\mathcal L_{self}$ denotes the self-supervised loss (see Section \ref{section3.4}).}
\label{fig:framework}
\end{figure*}

We here propose a Cross-Skeleton Interaction Graph Aggregation Network (CS-IGANet) to effectively learn abundant interaction relationships of mice, shown in Fig. \ref{fig:framework}. Our work is also based on GCN owing to its advantages in graph-structured data modelling. However, different from the above methods, we integrate GCN and Transformer to handle two aspects of social interactions (i.e., node-level and graph-level representation learning) by proposing a novel multi-level interaction module and a hierarchical interaction aggregation module.

Inspired by multi-view \cite{wang2020learning} and multi-scale \cite{liu2020disentangling,qi2023semantic} skeleton structures for individual action modelling, we propose a novel Cross-Skeleton Node-level Interaction (CS-NLI) module, shown in Fig. \ref{fig:framework}(b), to model the intra- (between keypoints of each mouse), inter- (between keypoints of different mice) and cross-skeleton (between keypoints of different skeletons) interactions of mice in a unified way. Unlike most existing methods \cite{wen2023interactive} for interactive action recognition, we consider the three types of interactions simultaneously to better learn multi-scale social interactions. Thus, we first introduce dense and sparse skeletons to describe multi-scale spatial structures of a mouse, shown in Fig. \ref{fig:framework}(a). Here, mouse skeleton refers to a list of keypoint connections \cite{lauer2022multi}. For each skeleton branch, a GCN-based module \cite{shi2019two} is first adopted to model the intra-skeleton interaction of each mouse, before we fuse dense geometric properties and velocity information. Afterwards, we model the social interactions of mice, where an adaptive inter-skeleton interaction matrix is formulated to integrate the motion information from two or more interactive mice. Similarly, we further explore the cross-skeleton interactions of mice. With the proposed multi-level interactions, our CS-NLI can discover abundant dynamic relations of social interactions, leading to more informative node-level representation.

% Then, a novel Cross-Skeleton Node-level Interaction (CS-NLI) module, shown in Fig. \ref{fig:framework}(b), is proposed to model the intra- (between keypoints of each mouse), inter- (between keypoints of different mice) and cross-skeleton (between keypoints of different skeletons) interactions in an unified way, allowing us to discover abundant dynamic relations of social interactions. For each skeleton branch, a GCN-based module \cite{shi2019two} is first adopted to model the intra-skeleton interaction of each mouse based on the extracted keypoints, before we fuse dense geometric properties and velocity information for multi-order dense information fusion. Afterwards, we model the social interactions of mice, where an adaptive inter-skeleton interaction matrix is formulated to integrate the motion information from two or more interactive mice. Similarly, we further explore the cross-skeleton interactions of these mice.

 Different from existing works \cite{shi2019two,liu2020disentangling,wen2023interactive} using GAP to directly generate graph-level representation, we propose to learn graph-level representation hierarchically while keeping crucial interaction information. This is achieved by a novel Interaction-Aware Transformer (IAT) guided by the proposed interaction-aware self-attention unit, shown in Fig. \ref{fig:framework}(c). The encoder aims at mining behaviour-related interaction saliency (i.e., conspicuous nodes) based on intra- and inter-skeleton interactions, where the node-level representation is used to generate multiple subgraphs and the last one denotes the graph-level representation of social behaviour.  Afterwards, such graph-level representations from different skeletons are integrated with representations generated by trivial pooling methods (e.g., average \cite{shi2019two}, max \cite{zhang2020semantics}) to enhance the graph-level representation. Moreover, a decoder is designed to adaptively update the node-level representation via the proposed interaction-aware self-attention.

% We also propose a novel Interaction-Aware Transformer (IAT) to dynamically learn graph-level representation of social behaviour, and update the node-level representation used as the input to the next layer, which is guided by the proposed interaction-aware self-attention unit, shown in Fig. \ref{fig:framework}(c). The encoder aims at mining behaviour-related interaction saliency (i.e., conspicuous nodes) based on intra- and inter-skeleton interactions, where the node-level representation is used to generate multiple subgraphs and the last one denotes the graph-level representation of social behaviour. Afterwards, such graph-level representations from different skeletons are integrated with representations generated by trivial pooling methods (e.g., average \cite{shi2019two}, max \cite{zhang2020semantics}) to enhance the graph-level representation. The decoder is designed so as to adaptively update the node-level representation via the proposed interaction-aware self-attention unit.

We believe that there exists meaningful similarity between the dense and sparse  skeletons that both describe spatial configurations of a mouse. To better preserve these attributes within the cross-skeleton pairwise nodes, we design an auxiliary self-supervised learning module. By jointly optimising the self-supervised objective function and the traditional classification loss function (i.e., cross-entropy loss), our proposed model can yield more discriminative representation.

% which further refines the classification results.

The main contributions can be summarised as follows:
\begin{itemize}
\item We propose a novel Cross-Skeleton Interaction Graph Aggregation Network (CS-IGANet) to learn mouse social behaviour representation, where dense and sparse skeletons cooperatively explore the spatio-temporal dynamics of social interactions.  

% To the best of our knowledge, we are the first attempt to quantify social behaviours for freely interacting mice by encoding skeleton information.

\item The proposed Cross-Skeleton Node-level Interaction (CS-NLI) module is able to engender powerful node-level representation by modelling multi-level interactions of mice, i.e., intra-, inter- and cross-skeleton interactions, where multi-order dense information are fused for inferring corresponding interaction patterns.

\item The proposed Interaction-Aware Transformer (IAT) allows for dynamic updating of graph- and node-level representation. This can be achieved by designing an encoder-decoder architecture, where the former hierarchically aggregates node-level representation for graph-level representation learning whilst the latter adaptively update node-level representation for extracting higher-level features.

\item We introduce an auxiliary self-supervised learning strategy to enable the proposed model to focus on the similarity between pairwise nodes from different skeletons, so as to enhance the representation ability of our model.

\end{itemize}

\section{Related Work}
% In this section, we review the established approaches related to our proposed network. Currently, the advanced methods for the modelling of skeleton data have been widely investigated in human action recognition domain. Therefore,  we first review the existing methods for pose-based mouse social behaviour recognition, followed by discussing the works used for skeleton-based human action recognition.

\subsection{Pose-based Mouse Social Behaviour Recognition}
Mouse behaviour recognition can be divided into two main categories: methods relying on RGB features and those utilising pose features. While most existing works \cite{jiang2018context,jiang2021muti,marks2022deep,camilleri2023mice} focus on extracting RGB features from videos, there is a limited exploration of mouse social behaviours through pose analysis. Giancardo et al. \cite{giancardo2013automatic} constructed a spatio-temporal feature vector composed of 13 measurements (e.g., relative position, shape and movement) based on the tracking results of the proposed tracker, and then applied random decision trees to classifying those extracted features. Similarly, Arac et al. \cite{arac2019deepbehavior} detected the nose, head, body and tail of each mouse using the standard YOLOv3 network, based on the extracting features such as distance between the body centers. However, these extracted features are shallow with limited spatio-temporal representation.

Thanks to pose estimation models \cite{Mathis2018, Graving2019,9492104}, people directly adopted the results of pose estimators to conduct downstream tasks such as behaviour recognition. Nilsson et al. \cite{nilsson2020simple} reported SimBa that analyses mouse social behaviours based on the pose estimation tracking results, where a random forest algorithm was leveraged to classify behavioural patterns. However, the 490 features (e.g. area of mouse convex hull, distance between part1 and part2) in their system are still shallow. Similar to SimBa, Segalin et al. \cite{segalin2021mouse} also introduced a system called MARS for the analysis of social behaviours, whereas 270 keypoint based spatio-temporal features were generated. However, these hand-crafted features cannot capture robust spatio-temporal relationships of keypoints, especially for complex social interactions.

\subsection{Skeleton-based Human Action Recognition}
\subsubsection{GCN-based methods}
Graph Convolutional Networks (GCNs) \cite{yan2018spatial,shi2019two,liu2020disentangling, plizzari2021skeleton, chen2021channel, hao2021hypergraph,9626596,heidari2021temporal} are prevalent for processing skeleton data due to their strong ability of capturing structural dependencies of joints.
The construction of GCNs generally follows two principles: spectral perspective \cite{sahbi2021learning,bruna2013spectral} and spatial perspective \cite{9412009,sahbi2010context}. 
Spectral methods leverage the eigenvalues and eigenvectors of the graph Laplace matrices, and they operate in the Fourier domain.  In contrast to spectral methods, spatial approaches, akin to traditional CNNs, perform convolutions directly in the spatial domain by aggregating information from local neighbourhoods. 
% Although spatial GCNs have generally outperformed spectral ones, their performance relies on attention matrices that capture contextual information and node-to-node relationships. To 
However, leveraging multi-head attention in GCNs improves accuracy but often leads to overparameterization and computational complexity \cite{sahbi2023phase}. To alleviate these issues, pruning-based methods \cite{sahbi2024tcmp,sahbi2022topologically,10448148} have emerged as mainstream, aiming to remove connections that have minimal impact on classification performance. Unlike these methods, our work integrates GCN and Transformer to handle two aspects of social interactions (i.e., node- and graph-level representation learning). This work follows the spatial methods.

Yan et al. \cite{yan2018spatial} exploited GCN for skeleton-based action recognition, and utilised the spatial temporal graph convolutional network (ST-GCN) to model the skeleton data as the graph structure.
However, it uses a fixed skeleton graph and represents only the physical structure of the human body. Shi et al. \cite{shi2019two} delineated a two-stream GCN model, i.e., 2s-AGCN to learn an adaptive graph where both the joint and bone information is explicitly utilised, significantly improving the model performance. Liu et al. \cite{liu2020disentangling} introduced a sophisticated feature extractor named MS-G3D, in which the disentangled multi-scale aggregator and G3D are used to eliminate redundant dependencies between neighbourhoods and model spatio-temporal information interaction, respectively.  Wang et al. \cite{wang2020learning} proposed a MV-IGNet network to formulate complementary action representations by adopting two pre-defined skeleton topologies. As we discussed above, it is difficult to determine a new and effective skeleton topology. Chen et al. \cite{chen2021channel} proposed to dynamically learn different topologies and effectively aggregate joint features in each channel. Chi et al. \cite{chi2022infogcn} designed a novel learning objective to learn compact latent representations. In \cite{hedegaard2023continual}, ST-GCN has been reformulated as a continual inference network, enabling online frame-by-frame predictions in a highly efficient manner. Although most of the aforementioned approaches have produced promising results in skeleton-based human action recognition, they mainly focus on single-object action without modelling the interactions between subjects, and hence lack the ability to generalise social representations.

\subsubsection{Transformer-based methods}
Transformer \cite{vaswani2017attention} using self-attention has also been applied to graph-structured data modelling due to its powerful ability of modelling long-range dependencies \cite{ yun2019graph, nguyen2019universal}. Recent studies have extended the Transformer model for skeleton-based action recognition \cite{plizzari2021skeleton,zhang2021stst,gedamu2023relation,huu2023step}. Plizzari et al. \cite{plizzari2021skeleton} proposed a spatio-temporal transformer network (i.e., ST-TR) for skeleton-based action recognition, where a spatial self-attention module was used to explore intra-frame interactions between different joints and a temporal self-attention module to model inter-frame correlations. Zhang et al. \cite{zhang2021stst} introduced a transformer network, where the spatial and temporal dimensions are parallelly separated. Nevertheless, this attention learning neglects the influence of different individual body joints on spatio-temporal action feature representations.
Huu et al. \cite{huu2023step} designed a hybrid architecture that combines GCN and Transformer to learn joint and body-part correlations using different cross-attention blocks. However,  these transformer-based networks are normally constrained by relatively high computational complexity.

\subsection{Interactive Action Recognition
} Recently proposed interactive action recognition methods \cite{zhu2021dyadic,li2022two,pang2022igformer} aim to capture spatio-temporal interactive features. Zhu et al. \cite{zhu2021dyadic} employed a GCN with separate graphs and proposed inter-body graph convolution with a dynamic relational adjacency matrix to capture interactions. Different from this work, Li et al. \cite{li2022two} introduced a novel two-person graph topology to represent inter-body and intra-body correlations. Other works \cite{pang2022igformer,wen2023interactive} have adopted the self-attention mechanism for human interaction modelling.  Pang et al. \cite{pang2022igformer} proposed to model the interactive relationship of subjects from both semantic and distance levels via an interaction graph Transformer. Wen et al. \cite{wen2023interactive} designed an interactive Spatio-temporal network to jointly model entity, temporal and spatial relations between interacting entities by fusing three-dimensional interactive spatio-temporal features. However, these approaches still face challenges in exploring mouse social interactions with ambiguous movement patterns.

\subsection{Mouse Pose Estimation}
Mouse pose estimation provides useful information for ethologically relevant behaviours. In recent years, deep learning based methods \cite{Mathis2018,Pereira2019,pereira2022sleap,9492104,marks2022deep} have been proposed for mouse or other animal pose estimation. Mathis et al. \cite{Mathis2018} proposed an animal pose estimation system called DeepLabCut, which adopts the feature detectors of DeeperCut with readout layers for markerless pose estimation. The system is trained with transfer learning, and it has been widely adopted in the behavioural research community. Similarly, LEAP \cite{Pereira2019} was developed to estimate poses in videos of individual mice and fruit flies, which provides a graphical interface for labelling body parts. However, its preprocessing is computationally expensive, thus limiting the application of their system in other environments. Pereira et al. \cite{pereira2022sleap} further designed a general framework called SLEAP for multi-animal pose estimation, which achieves comparable or improved accuracy compared to other systems for single-animal pose estimation with faster inference speed.

\section{Proposed Methods}
% \subsection{Preliminaries}
% \label{section3.1}
The skeleton sequence of $K$ mice with $T$ frames and $N$ keypoints can be represented as a spatio-temporal graph $\mathcal{G}=(\mathcal{V}, \mathcal{E},\mathcal{X})$. $\mathcal{V}=\left\{v_{t,n,k} \mid t,n,k \in \mathbb{Z}, 1\le t\le T, 1\le n\le N, 1\le k\le K\right\}$ is the set of all the nodes $v_{t,n,k}$ of the mouse skeleton graph, i.e., keypoints of the skeleton over all the time sequence. $\mathcal{E}$ represents the edge set consisting of two subsets, i.e., spatial topology $\mathcal{E_{S}}=\left\{(v_{t,n,k},v_{t,m,k}) \mid 1\le t\le T, 1\le n,m\le N, 1\le k\le K\right\}$ that describes the relationship between any pair of keypoints $(v_{n},v_{m})$ of mouse $k$ at time $t$, and temporal topology $\mathcal{E_{T}} = \left\{(v_{t,n,k},v_{t+1,n,k}) \mid, 1\le t\le T, 1\le n\le N, 1\le k\le K\right\}$ indicating the relationship between keypoints along consecutive time frames. $\mathcal{E_{S}}$ of each mouse at time $t$ can be formulated as an adjacency matrix $\textbf{A}\in \mathbb{R}^{N \times N}$ where initial element $a_{n,m}\in \left \{ 0,1 \right \} $ reflects the correlation strength between $v_{n}$ and $v_{m}$.  $\mathcal{X}=\left\{x_{t,n,k} \mid 1\le t\le T, 1\le n\le N, 1\le k\le K\right\}$ is a node features set, which is represented as a matrix $\mathbf{X}\in \mathbb{R}^{C \times T \times N \times K}$ where $x_{t, n,k}=\mathbf{X}(:,t,n,k) \in \mathbb{R}^{C}$ is the $C$ dimensional feature vector for node $v_{t,n,k}$. In this work, we focus on skeleton-based mouse social behaviour recognition in long videos. During training, we wish to obtain a continuous behaviour sequence by the sliding window over the long video, where each window centered at a certain frame only contains one specific behaviour \cite{jiang2021muti}. Hence, the behaviour sequence is represented as  $\mathbb{X}=\left[\mathbf{X}^{(1)},\mathbf{X}^{(2)} \ldots, \mathbf{X}^{(B)}\right] \in \mathbb{R}^{B \times C \times T \times N \times K} $, where $B$ is the total number of sliding windows and $\mathbf{X}^{(B)} \in \mathbb{R}^{C \times T \times N \times K} $  is the feature set of the $B$-th window in the long video. Consequently, given $\mathbb{X}$, we aim to learn a non-linear prediction function to model the relationship between a sequence of the predicted labels (i.e., $\mathbb{Y}=\left[\mathbf{Y}^{(1)},\mathbf{Y}^{(2)} \ldots, \mathbf{Y}^{(B)}\right] $) and $\mathbb{X}$. In experiments, following the standard formulations  \cite{yan2018spatial,shi2019two}, we reshape the input sequence to $\mathbb{X} \in \mathbb{R}^{K\cdot B \times C \times T \times N}$ by moving $K$ to the batch dimension. Normally, for each sliding window, one behaviour is described as $A$ and $\mathbf{X}  \in \mathbb{R}^{C \times T \times N}$, with $\mathbf{X}_{t} \in \mathbb{R}^{C \times N}$ being the node features at time $t$.

In this section, we will fully explain our proposed CS-NLI module that jointly models intra-, inter- and cross-skeleton interactions, our proposed IAT that dynamically creates graph-level representation and updates node-level representation, and the proposed auxiliary self-supervised learning strategy that encourages the proposed model to focus on the similarity between cross-skeleton pairwise nodes. The overview of our proposed framework is illustrated in Fig. \ref{fig:framework}.

\subsection{Cross-Skeleton Node-level Interaction}
\label{section3.2}

 % Mouse social behaviour recognition is non-trivial because it involves not only the individual behaviour but also the interactions between mice.

\begin{figure}
\begin{center}
\includegraphics[width=8.8cm]{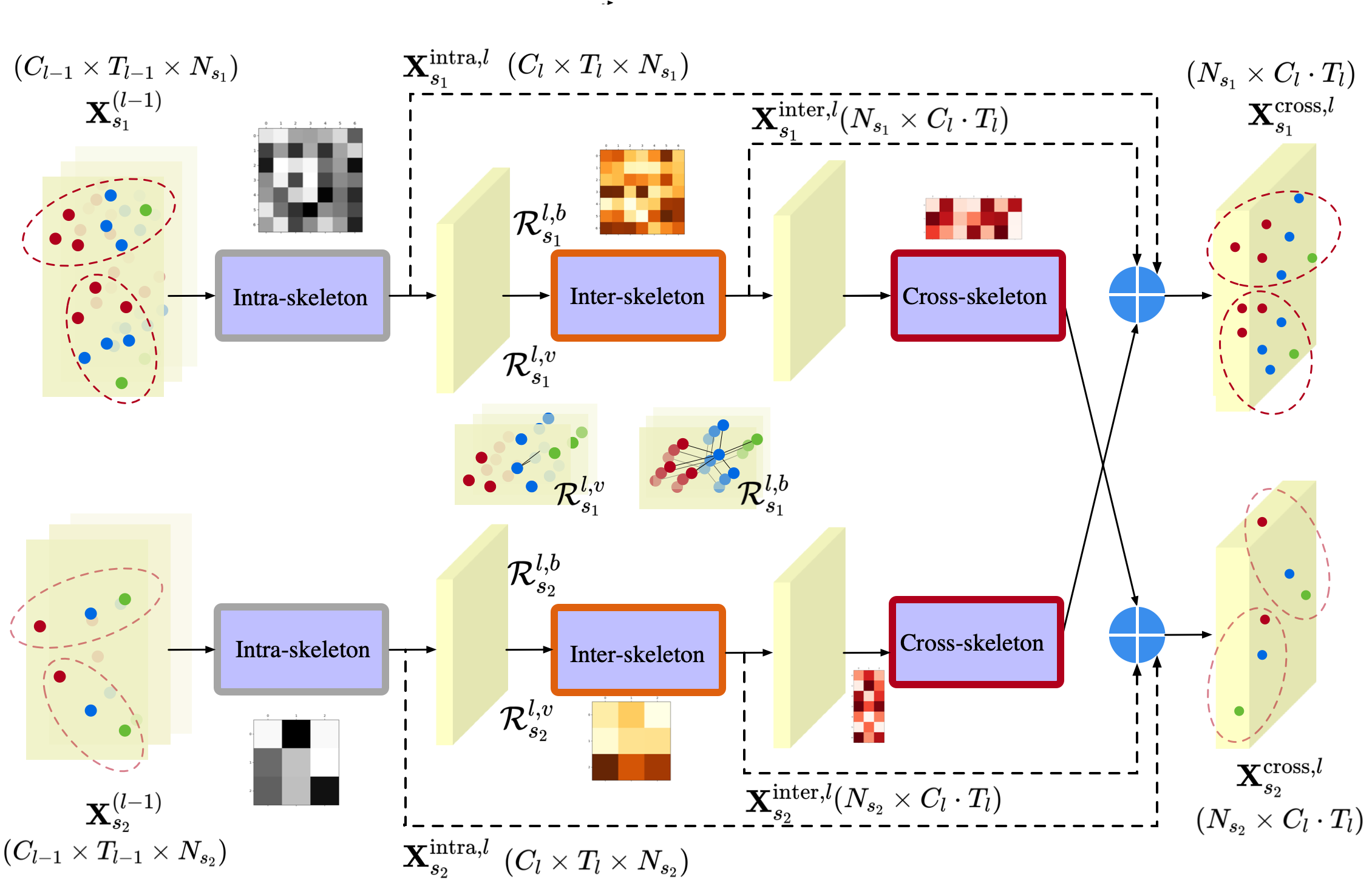}
\end{center}
\caption{Architecture of the Cross-Skeleton Node-level Interaction module (CS-NLI).} The inputs are node-level representations (circles with different colours) of the dense and sparse skeletons in the $(l-1)$-th layer, and each input contains the features of the mice. For each branch, $\mathbf{X}_{s_{i}} ^{\text{intra},l}$, $\mathbf{X}_{s_{i}} ^{\text{inter},l}$ and $\mathbf{X}_{s_{i}} ^{\text{cross},l}$ can be calculated by Eqs. (\ref{equation:gcn-tcn}), (\ref{equation:after_inter}) and (\ref{equation:after_cross-skeleton}), respectively.
This architecture is flexible and can be extended to deal with more skeletons and social behaviour recognition with more objects.
\label{fig:CS-NLI}
\end{figure}

% Mouse social behaviour recognition is non-trivial due to the intricate nature of not just individual behaviours but also the interactions of mice. The challenges arise from the fact that mouse social interactions exhibit ambiguous movement patterns, making it difficult to recognise and interpret specific behaviours. 
Unlike traditional behaviour recognition tasks that focus solely on individual behaviours \cite{jiang2018context}, understanding mouse social behaviour requires capturing the nuanced dynamics arising from the collective movements and interactions within a group of mice. Although the behavioural representation of each mouse on the both spatial and temporal domains can be interpreted by existing GCN-based network \cite{shi2019two, liu2020disentangling,wang2020learning}, they ignore the interaction between mice, which is crucial for fully learning the social behaviour representation. Therefore, in this section, we aim to explore the interaction between the keypoints of each mouse (i.e., intra-skeleton interaction) as well as interaction patterns between mice (i.e., inter-skeleton interaction) simultaneously. As aforementioned, we further model the spatio-temporal relationship between dense and sparse skeletons (i.e., cross-skeleton interaction) to learn skeleton-shared representations. The architecture of our proposed CS-NLI module is shown in Fig. \ref{fig:CS-NLI}.

\subsubsection{Intra-skeleton interaction modelling}
\label{section3.2.1}
Similar to \cite{wang2020learning, li2020dynamic}, we first construct $I$ types of skeleton sequences to learn behavioural information of mice. Following \cite{lauer2022multi}, we define the dense physical connections of all the keypoints to form the dense skeleton structure of each mouse, as shown in Fig. \ref{fig:framework}(a). Then, we further design a sparse structure where keypoints in the same body part are aggregated into one keypoint. The transition from dense skeleton structure to sparse skeleton structure is inspired by the established human body parts structure \cite{qi2023semantic,song2020stronger, huu2023step}. In these works, the average operation is normally utilised to aggregate neighbouring joints to obtain part-based skeleton structure. While acknowledging the structural differences between mouse and human bodies, we can apply a similar aggregation method to constructing the sparse skeleton structure consisting of three mouse parts, i.e., head, body and tail. The multi-scale skeleton graphs facilitate the extraction of behaviour-relevant features across different levels of granularity (
a comparison with CS-NLI using two dense graphs is provided in Tab. \ref{tab:CS-NLI_two_densegraphs}). This is because some behaviours such as ‘approach’ can be identified based on the movements of keypoints from the sparse skeleton without knowing the exact locations of each keypoint (e.g., left and right ears).

 To model the intra-skeleton interaction of mice (without loss of generality, we use two mice in this paper), we adopt the GCN-TCN structure shown in \cite{shi2019two} to encode the spatio-temporal representation (more details about this model can be found in Supplementary A). We adopt the standard GCN to extract spatial features from the structural node connections due to its flexibility on skeleton modelling. TCN is then used to extract temporal features from skeleton sequences. A residual connection is also added for both GCN and TCN. Mathematically, given the skeleton sequence $ \mathbf{X}_{s_{i}} ^{l}  \in \mathbb{R}^{C_{l} \times T_{l} \times N_{s_{i}}},\forall i \in \left \{ 1,2,\cdots ,I \right \} $, where $s_{i}$ and $N_{s_{i}}$ denote the $i$-th skeleton and the number of the nodes in this skeleton, we define such interaction as follows:
\begin{equation}\small
\begin{split}
\begin{aligned}
\mathbf{X}_{s_{i}} ^{(l+1)}&=\mathbf \Gamma (\Phi (\mathbf{X}_{s_{i}} ^{l}))+\mathbf{X}_{s_{i}} ^{l}
% &=\mathbf \Gamma(\sum_{k}^{K_{v}} \mathbf{W}_{k} \mathbf{X}_{s_{i}}^{l}\left (\mathbf{A}_{k}+\mathbf{B}_{k}+\mathbf{C}_{k}\right )+\mathbf{X}_{s_{i}} ^{l})+\mathbf{X}_{s_{i}} ^{l}
\end{aligned}
\label{equation:gcn-tcn}
\end{split}
\end{equation}
where $\Phi(\cdot)$ and $\Gamma(\cdot )$ represent spatial and temporal modelling, respectively. $l$ represents the $l$-th layer. The input $\mathbf{X}_{s_{i}} ^{l}$ in $\Phi(\cdot)$ is reshaped to $\mathbb{R}^{C_{l}\cdot T_{l} \times N_{s_{i}}}$ by assigning $T_{l}$ as the channel dimension, and is then projected back to $\mathbb{R}^{C_{l} \times T_{l} \times N_{s_{i}}}$ for temporal convolution. We can obtain the intra-skeleton interaction by stacking multiple residual GCN and TCN modules. The output of such interaction in the $l$-th layer of our network is represented as $\mathbf{X}_{s_{i}} ^{\text{intra}},l$.

\subsubsection{Inter-skeleton interaction modelling}
\label{section3.2.2}
Based on the high-level features extracted using a sequence of residual GCN and TCN modules defined in Eq. (\ref{equation:gcn-tcn}), we then explore the interaction pattern between mice where the inter-skeleton interaction matrix, i.e., $\mathbf{A}_{k_{i}\rightarrow k_{j} }^{l}$, needs to be derived (for two mice, we have $\mathbf{A}_{k_{1}\rightarrow k_{2} }^{l}$ and $\mathbf{A}_{k_{2}\rightarrow k_{1} }^{l} $). We first explicitly embed both geometric distance and velocity information into the representation encoded by keypoint information because they carry behaviour-related features \cite{wang2020learning,shi2019two,song2020stronger}. In particular, mouse body is highly deformable and most mouse behaviours have a relatively short duration. Unlike these approaches, we here extract dense geometric distance (Eq. (\ref{equation:bone})) and velocity information (Eq. (\ref{equation:velicity})) simultaneously. For the node $v_{m}$ at the $l$-th layer of our network, we consider the relative positions between it and all the remaining nodes to construct a dense geometric distance set $\mathcal{R}_{s_{i}}^{l,b}=\left\{r_{s_{i}}^{l,b}(n,m) \mid n=1,2, \cdots, N_{s_{i}}\right\}$, where
\begin{equation}
\begin{split}
r_{s_{i}}^{l,b}(n,m)=\mathcal R_{s_{i},n,m}^{l,b}  =\mathbf{X}_{s_{i},n} ^{l}-\mathbf{X}_{s_{i},m} ^{l}
\label{equation:bone}
\end{split}
\end{equation}
$\mathbf{X}_{s_{i},m} ^{l}=\mathbf{X}_{s_{i}}^{l}(:,:,m) \in \mathbb{R}_{s_{i}}^{C_{l} \times T_{l}} $ is the feature of node $v_{m}$ across the temporal domain. Similarly, we produce a dense velocity set of time $t$, i.e., $\mathcal{R}^{l,m}=\left\{r_{s_{i}}^{l,m}(p,t) \mid p=1,2, \cdots, T_{l}\right\}$ with the following definition:
\begin{equation}
\begin{split}
r_{s_{i}}^{l,v}(p,t)=\mathcal R_{s_{i},p,t}^{l,v}  =\mathbf{X}_{s_{i},p} ^{l}-\mathbf{X}_{s_{i},t} ^{l}
\label{equation:velicity}
\end{split}
\end{equation}
where $\mathbf{X}_{s_{i},t} ^{l}=\mathbf{X}_{s_{i}} ^{l}(:,t,:) \in \mathbb{R}^{C_{l}\times N_{s_{i}}}$ represents the feature of all the nodes at time $t$.

% Before we aggregate the multi-order information, a TCN is used to model temporal relations, i.e., $\overline{\mathbf{X}}_{s_{i}} ^{l}=\mathbf \Gamma ({\mathbf{X}}_{s_{i}} ^{l}) $, where the feature dimensions on the space and time are compressed to $\overline{C}_{l}$ and $\overline{T}_{l}$ respectively for faster inference. Then we can get the corresponding geometric distance information, i.e., $ \overline{\mathcal R}_{s_{i},n,m}^{l,b}$ (for node  $v_{n}$ and $v_{m}$) and velocity information, i.e., $\overline{\mathcal R}_{s_{i},p,t}^{l,v}$ (for time $p$ and $t$) according to Eqs. (\ref{equation:bone}) and (\ref{equation:velicity}), respectively.

We integrate the features of node $v_{m}$ over the temporal domain (i.e., $\mathbf{X}_{s_{i},m} ^{l}$ ) with its dense geometric distance, and the features of all the nodes at time $t$ (i.e., $\mathbf{X}_{s_{i},t} ^{l}$ ) with its dense velocity. For the former, $\mathbf{X}_{s_{i},m} ^{l}$ and one of the elements of set $\mathcal{R}_{s_{i}}^{l,b}$ are fused by concatenation, followed by performing dimensionality reduction on features. All the features are then stacked together, and we also add a residual connection in order to obtain the features of node $v_{m}$ (i.e., $\mathbf{H}_{s_{i},m} ^{l,b}$), fusing the dense geometric distance. Similarly, we can obtain the features of all the nodes at time $t$  (i.e., $\mathbf{H}_{s_{i},t} ^{l,v}$), using the dense velocity information. We have the following expression:
\begin{equation}
\begin{split}
\mathbf{H}_{s_{i},m} ^{l,b}=\bigg(\sum_{n=1}^{N_{s_{i}} }g_{s_{i}}^{b}([\mathbf{X}_{s_{i},m} ^{l};\mathcal{ R}_{s_{i},n,m}^{l,b}])\bigg)+\mathbf{X}_{s_{i},m} ^{l}
\label{equation:multi-order1}
\end{split}
\end{equation}

\begin{equation}
\begin{split}
\mathbf{H}_{s_{i},t} ^{l,v}=\bigg(\sum_{p=1}^{T_{l} }g_{s_{i}}^{v}([\mathbf{X}_{s_{i},t} ^{l};\mathcal{ R}_{s_{i},p,t}^{l,v}])\bigg)+\mathbf{X}_{s_{i},t} ^{l}
\label{equation:multi-order2}
\end{split}
\end{equation}
where $\mathbf{X}_{s_{i},m} ^{l}$ and $\mathbf{X}_{s_{i},t} ^{l}$ are reshaped to $\mathbb{R}^{{C}_{l}\cdot {T}_{l}} $ and $\mathbb{R}^{ {C}_{l}\cdot N_{s_{i}}} $, respectively. $g_{s_{i}}^{b}(\cdot)$ and $g_{s_{i}}^{v}(\cdot)$ denote the Multi-Layer Perceptrons (MLPs). $[;]$ represents the concatenation operation. We then obtain the multi-order dense information embedded with keypoints, geometric distance as well as velocity by fusing $\mathbf{H}_{s_{i},m} ^{l,b}$ and $\mathbf{H}_{s_{i},t} ^{l,v}$, as follows:
\begin{equation}
\begin{split}
\mathbf{H}_{s_{i},m} ^{l}=f_{s_{i}}^{}[ \mathbf{H}_{s_{i},m} ^{l,b};\varepsilon (\mathbf{H}_{s_{i},t} ^{l,v})] \in \mathbb{R}^{{C}_{l}\cdot  {T}_{l}}
\label{equation:multi-order3}
\end{split}
\end{equation}
where $f_{s_{i}}(\cdot)$ denotes the MLPs. $\varepsilon(\cdot)$ reshapes $\mathbf{H}_{s_{i}} ^{l,v}$ from $\mathbb{R}^{ {T}_{l} \times {C}_{l}\cdot N_{s_{i}}}$ to $\mathbb{R}^{N_{s_{i}} \times  {C}_{l}\cdot  {T}_{l}}$, where the $N_{s_{i}}$ dimension has been moved to the first position so that the geometric distance and velocity can be fused by the concatenation operation. Here, the information of each mouse can be represented as $(\mathbf{H}_{s_{i},m} ^{l})_{k_{1}}$ and $(\mathbf{H}_{s_{i},m} ^{l})_{k_{2}}$.

% In experiments, $\mathbf{H}_{s_{i},m} ^{l}$ contains the features of two mice, which is decomposed into $(\mathbf{H}_{s_{i},m} ^{l})_{k_{1}}$ and $(\mathbf{H}_{s_{i},m} ^{l})_{k_{2}}$.

Given the aggregated representation of two mice $\mathbf{H}_{s_{i}} ^{l}$, our target is to learn an inter-skeleton interaction pattern. Therefore, the interaction between node $v_{m}$ of mouse $k_{1}$ with representation $(\mathbf{H}_{s_{i},m} ^{l})_{k_{1}}$ and node $v_{n}$ of mouse $k_{2}$ with representation $(\mathbf{H}_{s_{i},n} ^{l})_{k_{2}}$ can be expressed as:
\begin{equation}
\begin{split}
\begin{aligned}
\mathbf{E}_{k_{1}\to k_{2}}^{l}(m, n)=&\sigma\left(\overrightarrow{\mathbf{\varpi }}\left[(\mathbf{H}_{s_{i},m} ^{l})_{k_{1}}; (\mathbf{H}_{s_{i},n} ^{l})_{k_{2}}\right]^{\top}\right)+\\
&\beta \cdot \mathbf{\Lambda}_{k_{1}\to k_{2}}(m, n)
\end{aligned}
\label{equation:inter-skeleton}
\end{split}
\end{equation}
where $\sigma$ is the activation function as ReLU. $\overrightarrow{\mathbf{\varpi }} \in \mathbb{R}^{1 \times 2 {C}_{l}\cdot  {T}_{l}}$ denotes a learnable weight vector. $\mathbf{\Lambda}_{k_{1}\to k_{2}}$ denotes the pre-defined physical connections describing the inter-skeleton interaction between mice, where $\mathbf{\Lambda}_{k_{1}\to k_{2}}=1$ if node $v_{m}$ of mouse $k_1$ and $v_{n}$ of mouse $k_2$ is connected. Specifically, these connections link corresponding keypoints of different mice, such as the nose of mouse $k_1$ corresponding to the nose of mouse $k_2$.
Similar to AGCN \cite{shi2019two}, our inter-skeleton interaction matrix is composed of both fixed and learnable matrices. $\mathbf{\Lambda}_{k_{1}\to k_{2}}$ is fixed, and it is combined with a learnable matrix to generate the final interaction matrix $\mathbf{E}_{k_{1}\to k_{2}}^{l}$.
In particular, we introduce a trade-off parameter $\beta $ to balance the potential effect of the pre-defined interaction pattern. Thus, the generated $\mathbf{E}_{k_{1}\to k_{2}}^{l}(m, n)$ represents the correlation degree between the two nodes, and it is also dynamically updated to learn behaviour-specific inter-skeleton interaction. Besides, we normalise the results in Eq. (\ref{equation:inter-skeleton}) by the $Softmax$ function to allow the correlation degree to be comparable, as follows:
\begin{equation}
\begin{split}
\begin{aligned}
\mathbf{A}_{k_{1}\to k_{2}}^{l}(m, n)=\frac{\exp \left(\mathbf{E}_{k_{1} \to k_{2}}^{l}(m, n)\right)}{\sum_{v=1}^{N_{s_{i}}}\exp \left(\mathbf{E}_{k_{1} \to k_{2}}^{l}(m, v)\right)}
\end{aligned}
\label{equation:inter_A}
\end{split}
\end{equation}

In this paper, we design a bidirectional interaction model, i.e., impact of mouse $k_{1}$ on $k_{2}$ and that of mouse $k_{2}$ on $k_{1}$, to fully explore potential inter-skeleton interaction. Thus, the interaction from $k_{2}$ to $k_{1}$, i.e., $\mathbf{A}_{k_{2}\to k_{1}}^{l}$, can also be inferred using the same method shown in Eqs. (\ref{equation:inter-skeleton}) and (\ref{equation:inter_A}). Afterwards, we generate the node-level representation integrated with the inter-skeleton interaction. Given the spatio-temporal representation of a mouse, after intra-skeleton interaction modelling, the behavioural representation of another mouse is updated as:
\begin{equation}
\begin{split}
\begin{aligned}
(\mathbf {X}_{s_{i}}^{\text{inter},l})_{k_{1}} & =  \mathbf{A}_{k_{2}\to k_{1}}^{l} (\mathbf{X}_{s_{i}}^{\text{intra},l})_{k_{2}}\mathbf W_{s_{i},k_{2}\to k_{1}}^{l}  + (\mathbf X_{s_{i}}^{l})_{k_{1}}\\
(\mathbf {X}_{s_{i}}^{\text{inter},l})_{k_{2}} & =  \mathbf{A}_{k_{1}\to k_{2}}^{l} (\mathbf{X}_{s_{i}}^{\text{intra},l})_{k_{1}}\mathbf W_{s_{i},k_{1}\to k_{2}}^{l}  + (\mathbf X_{s_{i}}^{l})_{k_{2}}
\end{aligned}
\label{equation:after_inter1}
\end{split}
\end{equation}
where $\mathbf W_{s_{i},k_{2}\to k_{1}}^{l}$ and $\mathbf W_{s_{i},k_{1}\to k_{2}}^{l} \in  \mathbb{R}^{  {C}_{l}\cdot {T}_{l} \times   {C}_{l}\cdot {T}_{l}}$ are trainable weight matrices. Then, we compose the representations of the mice to generate the node-level representation embedded with inter-skeleton interaction $\mathbf X_{s_{i}}^{\text{inter},l}$ by concatenation on the batch dimension, as shown in Eq. (\ref{equation:after_inter}):
\begin{equation}\small
\begin{split}
\begin{aligned}
\mathbf X_{s_{i}}^{\text{inter},l}=\Psi(\mathbf X_{s_{i}}^{l})=[(\mathbf X_{s_{i}}^{\text{inter},l})_{k_{1}};(\mathbf X_{s_{i}}^{\text{inter},l})_{k_{2}}] \in \mathbb{R}^{   N_{s_{i}} \times C_{l}T_{l} }
\end{aligned}
\label{equation:after_inter}
\end{split}
\end{equation}

% We exchange the order of $k_{1}$ and $k_{2}$ to generate $\widetilde{\mathbf X}_{s_{iJ}}^{inter,l}$,  $\mathbf X_{s_{i}}^{inter,l}$,

\subsubsection{Cross-Skeleton interaction modelling}
\label{section3.2.3}
In this subsection, we aim to model the cross-skeleton interaction within the same mouse, and that between different mice at the same time.
% Additionally, the inference of the inter- and cross-skeleton interaction patterns is based on the multi-order dense information, which facilitates complex social interaction learning to some extent.
Based on $\mathbf X_{s_{i}}^{\text{inter},l}$ shown in Eq. (\ref{equation:after_inter}), we first learn skeleton-shared representation within the same mouse. Similar to the inter-skeleton interaction, the relation degree between the $n$-th keypoint of one mouse and the $m$-th body part of the same mouse needs to be derived. Thus, to integrate information from $s_{1}$ (i.e., dense skeleton) to $s_{2}$ (i.e., sparse skeleton), we rewrite Eq. (\ref{equation:inter-skeleton}) by combining  Eq. (\ref{equation:inter_A}) as follows:
\begin{equation}
\begin{split}
\begin{aligned}
\mathbf{A}_{s_{1}\to s_{2}}^{l}(m, n)=&Softmax \bigg(\sigma \Big(\overrightarrow{\rho } [\mathbf{H}_{s_{2},m} ^{l} ; \mathbf{H}_{s_{1}, n} ^{l}]^{\top}\Big)  +\\
&\beta   \cdot \mathbf{\Lambda }_{s_{1}\to s_{2}}(m, n)\bigg)
\end{aligned}
\label{equation:cross-skeleton}
\end{split}
\end{equation}
where $\overrightarrow{\mathbf{\rho }} \in \mathbb{R}^{1 \times 2 {C}_{l}\cdot  {T}_{l}}$ denotes a learnable weight vector. $\mathbf{\Lambda}_{s_{1}\to s_{2}}$ is the predefined connections across the overall skeletons  (e.g, nose, left ear and right ear in $s_1$ correspond to head in $s_2$). Similarly, we model the cross-skeleton interaction of different mice. We exchange the orders of $k_{1}$ and $k_{2}$ in Eq. (\ref{equation:after_inter}) to generate $\widetilde{\mathbf X}_{s_{i}}^{\text{inter},l}$, and the order of the mouse in $\mathbf{H}_{s_{1}, n} ^{l}$ used in Eq. (\ref{equation:cross-skeleton}) to yield $\widetilde{\mathbf{A}}_{s_{1}\to s_{2}}^{l}(m, n)$.

Finally, we have the corresponding node-level representation of $s_{2}$ (i.e., $\mathbf X_{s_{2}}^{\text{cross},l} \in \mathbb{R}^{N_{s_{2}} \times {C}_{l}\cdot{T}_{l}}$ ) by fusing the information from $s_{1}$, including four parts, i.e., the initial intra-skeleton interaction information, inter-skeleton interaction information, cross-skeleton interaction information of the same mouse and cross-skeleton interaction information of different mice. Hence, we can have:
\begin{equation}
\begin{split}
\begin{aligned}
\mathbf X_{s_{2}}^{\text{cross},l} & = X_{s_{2}}^{\text{intra},l}+\mathbf X_{s_{2}}^{\text{inter},l}+\mathbf{A}_{s_{1}\to s_{2}}^{l}\mathbf X_{s_{1}}^{\text{inter},l}\mathbf W_{s_{1}\to s_{2}}^{l}  +
\\&\widetilde{\mathbf{A}}_{s_{1}\to s_{2}}^{l}\widetilde{\mathbf X}_{s_{1}}^{\text{inter},l}\widetilde{\mathbf W}_{s_{1}\to s_{2}}^{l}
\end{aligned}
\label{equation:after_cross-skeleton}
\end{split}
\end{equation}
where$\mathbf W_{s_{1}\to s_{2}}^{l}$, $\widetilde{\mathbf W}_{s_{1}\to s_{2}}^{l} \in  \mathbb{R}^{  {C}_{l}\cdot {T}_{l} \times   {C}_{l}\cdot {T}_{l}}$ are  trainable weight matrices. The fusion from  $s_{2}$ to $s_{1}$ can also be made using the same way as mentioned above.

\subsection{Interaction-Aware Transformer}
\label{section3.3}
\begin{figure}
\begin{center}
\includegraphics[width=8.4cm]{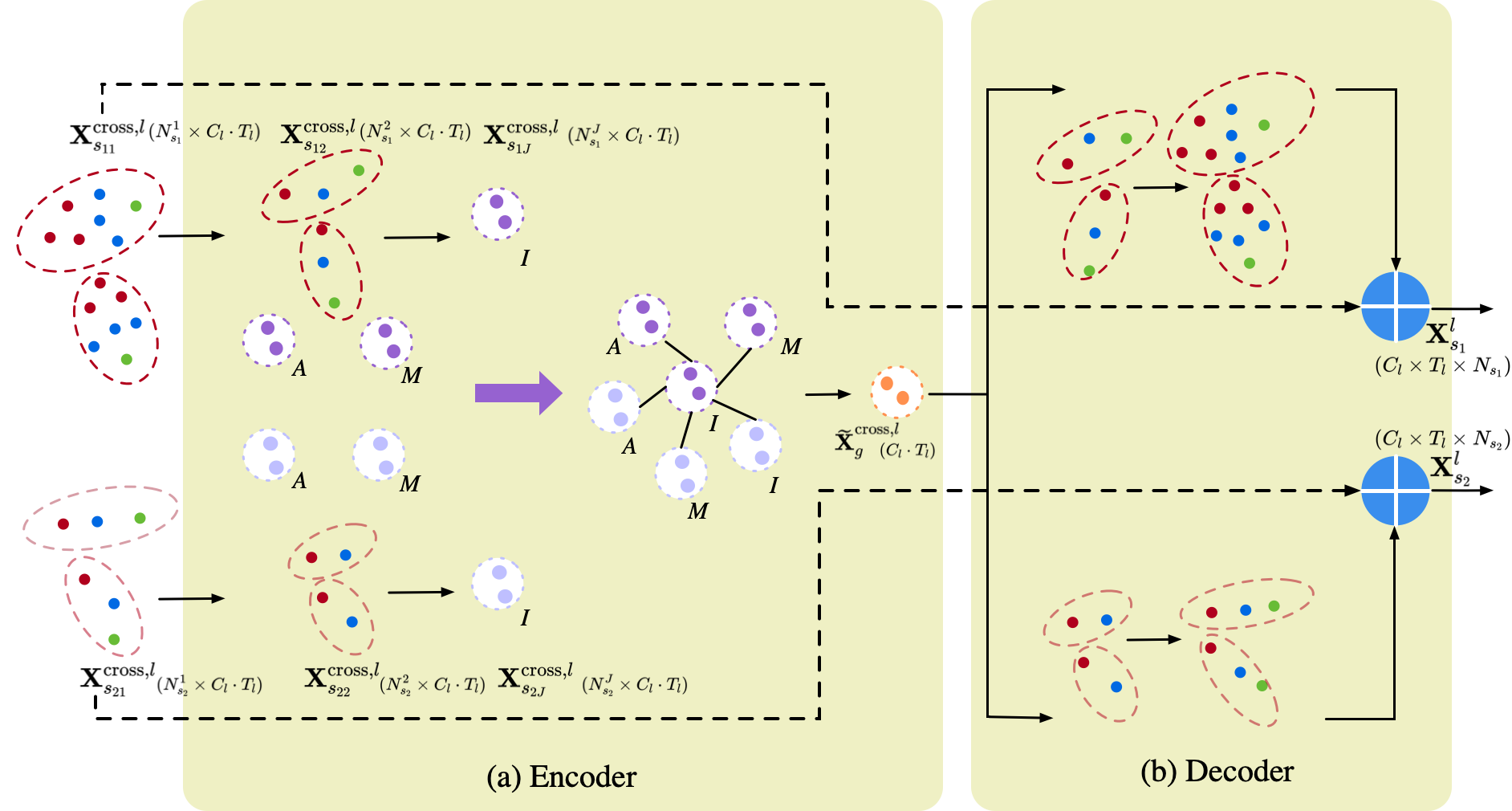}
\end{center}
\caption{Architecture of the Interaction-Aware Transformer (IAT). (a) The inputs of the encoder are node-level representations from the CS-NLI module, i.e., the first subgraph in the encoder. For clear illustration, we draw several circles with different colours to show the dynamic process of graph-level representation learning. (b) The outputs of the decoder are the node-level representations from two skeletons. }
\label{fig:IAT}
\end{figure}

In this section, we aim to generate graph-level representation of mouse social behaviour for classification from the CS-NLI module reported in Section \ref{section3.2}, and further update node-level representation used as the input to the next layer for capturing higher-level features. The architecture of our proposed IAT is shown in Fig. \ref{fig:IAT}.

\subsubsection{Encoder}
\label{section3.3.1}
As aforementioned in Section \ref{section1}, significant interaction information of mice may be lost if we adopt the pooling operation such as the global average pooling to produce graph-level representation.
Hence, to learn a discriminative graph-level representation, we design a novel Interaction-Aware Transformer based on the self-attention mechanism \cite{vaswani2017attention, dehghani2018universal}. Given the node-level representation of skeleton graph $s_{i}$ on the $l$-th layer (i.e., $\mathbf X_{s_{i}}^{\text{cross},l} \in \mathbb{R}^{N_{s_{i}} \times {C}_{l}\cdot{T}_{l}}$), we sequentially generate $J$ subgraphs, i.e., $\mathbf X_{s_{ij}}^{\text{cross},l} \in \mathbb{R}^{N_{s_{i}}^{j} \times {C}_{l}\cdot{T}_{l}}$, with $N_{s_{i}}^{j}, \forall j \in \left \{ 1,2,\dots,J  \right \}$ nodes. Inspired by the universal transformer model \cite{dehghani2018universal}, we design an interaction-aware self-attention mechanism using a self-attention block,  followed by a recurrent transition for hierarchical spatio-temporal representation learning. Regarding node $v_{m_{j}}\in \mathcal N_{s_{i}}^{j}  = \left \{ 1,2,\dots,N_{s_{i}}^{j}  \right \} $ in  the $j$-th subgraph (the first subgraph represents the node-level representation generated by the CS-NLI module), we have:
\begin{equation}\small
\begin{split}
\begin{aligned}
\mathbf H_{s_{ij+1},m_{j+1}}^{\text{cross},l}  = LN(\mathbf Q_{s_{ij+1},m_{j+1}}^{l} + \mathbf \Upsilon _{m_{j+1}}(\mathbf X_{s_{ij},m_{j}}^{\text{cross},l}))
\end{aligned}
\label{equation: selfatt}
\end{split}
\end{equation}
\begin{equation}\small
\begin{split}
\begin{aligned}
\mathbf X_{s_{ij+1},m_{j+1}}^{\text{cross},l} =LN(\mathbf H_{s_{ij+1},m_{j+1}}^{\text{cross},l}+\mathbf \Gamma (\mathbf H_{s_{ij+1},m_{j+1}}^{\text{cross},l}) )
\end{aligned}
\label{equation: selfatt2}
\end{split}
\end{equation}
where $LN(\cdot)$ is to normalise the inputs across the entire feature dimensions. The transition function shown in Eq. (\ref{equation: selfatt2}) is defined as a TCN that models the temporal relations between nodes. $\mathbf \Upsilon( \cdot)$ denotes the self-attention network to dynamically learn the graph-level representation, which can be formulated as:
\begin{equation}
\begin{split}
\begin{aligned}
\mathbf \Upsilon _{m_{j+1}}(\mathbf X_{s_{ij+1},m_{j}}^{\text{cross},l})=\sum_{m_{j}\in \mathcal N_{s_{i}}^{j}}\alpha _{m_{j+1},m_{j}}^{l}\mathbf V_{s_{ij},m_{j}}^{l}
\end{aligned}
\label{equation: selfatt3}
\end{split}
\end{equation}
where $\alpha _{m_{j+1},m_{j}}^{l}$ represents the strength of the correlations between nodes $v_{m_{j+1}}$ and $v_{m_{j}}$, based on the query and key vectors. $\mathbf V_{s_{ij},m_{j}}^{l}$ is the value vector of $m_{j}$, and the score $\alpha _{m_{j+1},m_{j}}^{l}$ is used to weight each node's key vector.

The query, key and value vectors in the Transformer architecture are used to establish a self-attention mechanism \cite{vaswani2017attention}. Different from most self-attention methods of applying linear transformations \cite{dehghani2018universal, lee2019set} or convolution \cite{plizzari2021skeleton} to the node features, we propose an interaction-aware self-attention approach to construct the vectors, based on the structural intra- and inter-skeleton interactions. Particularly, we focus on the behaviour-related interaction saliency by condensing $N_{s_{i}}^{j}$ nodes into a sub-graph with $N_{s_{i}}^{j+1}$ nodes. Hence, the query vector is defined as:
\begin{equation}
\begin{split}
\begin{aligned}
\mathbf Q_{s_{ij},m_{j+1}}^{l}=\sum_{m_{j} \in N_{s_{i}}^{j} } \mathbf W_{s_{ij}}^{l}(m_{j+1},m_{j})\mathbf X_{s_{ij},m_{j}}^{\text{cross},l}
\end{aligned}
\label{equation: selfatt_Q}
\end{split}
\end{equation}
where $\mathbf W_{s_{ij}}^{l}(m_{j+1},m_{j})$ are the elements of a trainable matrix $\mathbf W_{s_{ij}}^{l} \in  \mathbb{R}^{N_{s_{i}}^{j+1} \times N_{s_{i}}^{j} } $ at the $m_{j+1}$-th row and $m_{j}$-th column. The output $\mathbf Q_{s_{ij},m_{j+1}}^{l}$ denotes the feature vector of node $v_{m_{j+1}}$. We next compute the key and value vectors for node $v_{m_{j}}$, which are jointly encoded by different interaction patterns, i.e., intra- and inter-skeleton interactions. The key vector $\mathbf K_{s_{ij},m_{j}}^{l}$ can be computed as follows:
\begin{equation}
\begin{split}
\begin{aligned}
\mathbf K_{s_{ij},m_{j}}^{l}=Conv_{1\times 1} \Big([\Phi(\mathbf X_{s_{ij},m_{j}}^{\text{cross},l});\Psi (\mathbf X_{s_{ij},m_{j}}^{\text{cross},l}) ] \Big)
\end{aligned}
\label{equation: selfatt_k}
\end{split}
\end{equation}
where $\Phi(\mathbf X_{s_{ij},m_{j}}^{\text{cross},l})$  representing the intra-skeleton interaction on the spatial domain for node $v_{m_{j}}$ can be formed using Eq. (\ref{equation:gcn-tcn}). $\Psi (\mathbf X_{s_{ij},m_{j}}^{\text{cross},l})$ denotes the corresponding inter-skeleton interaction that is calculated by combining Eqs. (\ref{equation:after_inter1}) and (\ref{equation:after_inter}). The concatenation is performed on the channel dimension, and $Conv_{1\times 1}(\cdot)$ denotes $1 \times 1$ convolution to reduce the channel dimension.

Similarly, we formulate the value vector $\mathbf V_{s_{ij},m_{j}}^{l}$ in Eq. (\ref{equation: selfatt3}) according to Eq. (\ref{equation: selfatt_k}). Then, the attention weight $\alpha _{m_{j+1},m_{j}}^{l}$ can be defined by applying the $Softmax$ function to the scaled dot products \cite{vaswani2017attention} between $m_{j+1}$ and  $m_{j}$:
\begin{equation}
\begin{split}
\begin{aligned}
\alpha _{m_{j+1},m_{j}}^{l}=Softmax_{m_{j}}(\frac{\mathbf Q_{s_{ij+1},m_{j+1}}^{l}(\mathbf K_{s_{ij},m_{j}}^{l})\top   }{\sqrt{C_{l}T_{l}} } )
\end{aligned}
\label{equation: selfatt_alph}
\end{split}
\end{equation}

To this end, we can hierarchically generate multiple subgraphs through Eqs. (\ref{equation: selfatt}) and (\ref{equation: selfatt2}), in which the final one, i.e., $\mathbf X_{s_{iJ},m_{J}}^{\text{cross},l}$ denotes the behaviour-related graph-level representation with $N_{s_{i}}^{J}=1$.

To generate the graph-level representation from the skeleton graph, GAP \cite{shi2019two} or max pooling \cite{zhang2020semantics} has been used to merge the information of all the keypoints or frames. Intuitively, different graph-level representations carry different semantic information describing social interactions. Thus, we also explore the relations between different graph-level representations by using our proposed interaction-aware self-attention defined in Eqs. (\ref{equation: selfatt}) and (\ref{equation: selfatt2}) to enhance the graph-level representation. More details can be found in Supplementary A.

\subsubsection{Decoder}
\label{section3.3.2}

In most prior works like \cite{yan2018spatial,shi2019two}, the node-level representation is directly passed between blocks in a GCN-TCN architecture. In contrast, we add a decoder at the end of the encoder to update node-level representation before passing it to the next layer. This update is supported by our interaction-aware self-attention mechanism presented in Section \ref{section3.3.1}. Supplementary A provides more explanations and our proposed IAT is summarised in Algorithm \ref{Alg:IAT}.

\subsection{Self-supervision for Cross-Skeleton Node Similarity Learning}
\label{section3.4}
% Self-supervised learning has been applied to image \cite{pan2020self}, video \cite{wang2021temporal} and graph \cite{jin2020self} domains by generating necessary supervisory signals (i.e., pseudo labels) which are derived from data itself. In this work, 

As aforementioned, there potentially exists important similarity between the two skeletons (dense and sparse skeletons) because they describe the spatial structure of the mouse from different scales. In other words, the similarity is naturally embedded into the node-level representations of the two skeletons, which may play a crucial role in social behaviour representation learning. Inspired by the attribute based self-supervision \cite{jin2020self} on graphs, we design an auxiliary self-supervised learning task to better preserve these attributes between the cross-skeleton pairwise nodes.

For the $i$-th sliding window, given the initial spatial-temporal feature of the node $v_{m}$ in skeleton $s_{1}$ (i.e., $\mathbf X_{s_{1},m}^{(i)}$) and that of $v_{n}$ in skeleton $s_{2}$ (i.e., $\mathbf X_{s_{2},n}^{(i)}$), we first compute the node feature similarity between them according to the cosine similarity:
\begin{equation}
\begin{split}
\begin{aligned}
S_{mn}^{(i)}=\frac{\mathbf X_{s_{1},m}^{(i)} \cdot\mathbf X_{s_{2},n}^{(i)}}{max \left(\left\|\mathbf X_{s_{1},m}^{(i)}\right\|_{2} \cdot\left\|\mathbf X_{s_{2},n}^{(i)}\right\|_{2}, \epsilon\right)}
\end{aligned}
\label{equation: selfsuper}
\end{split}
\end{equation}
where $\epsilon$ is a small constant avoiding divide-by-zero. Then, the self-supervised learning task can be formulated as a regression problem and the corresponding loss can be defined as:
\begin{equation}
\begin{split}
\begin{aligned}
\mathcal{L}_{\text {self}}=&\frac{1}{B} \sum_{i=1}^{B}\bigg (\frac{1}{|\mathcal{P}|} \sum_{\left(v_{m}, v_{n}\right) \in \mathcal{P}}
\\
&\left\|f_{s}\left(\mathbf{X}_{s_{1},m}^{(i),l}-\mathbf{X}_{s_{2},n}^{(i),l}\right)-\mathbf{S}_{mn}^{(i)}\right\|^{2} \bigg)
\end{aligned}
\label{equation: selfsuperloss}
\end{split}
\end{equation}
where $\mathcal{P}$ denotes the set of node pairs consisting of nodes from different skeletons, and $|\mathcal{P}|$ is the number of the nodes. $f_{s}(\cdot)$ is a fully connected layer with the output dimension of 1. $\mathbf{X}_{s_{1},m}^{(i),l}$ is the node-level representation of node $v_{m}$ in the $l$-th layer of our network.

Finally, we can obtain the overall behaviour recognition loss by combining the self-supervised loss defined in Eq. (\ref{equation: selfsuperloss}) and cross-entropy loss $\mathcal{L}_{\text {class }}$ (see Supplementary A), which is defined as $\mathcal{L}=\mathcal{L}_{\text {class }}+\lambda \mathcal{L}_{\text {self }}$, where $\lambda$ is a hyper-parameter adjusting the contribution of the self-supervised loss. By jointly optimising the self-supervised objective function and the traditional classification loss function, our proposed model can lead to more discriminative representation.

\section{Experiments}
\label{section4}

\subsection{Datasets and Experimental Setup}
\label{section4.1}
\subsubsection{CRIM13-Skeleton Dataset}
\label{section4.1.1}
In this paper, we validate the proposed framework using videos of two mice. The Caltech Resident-Intruder Mouse dataset \cite{burgos2012social} (CRIM13) consists of 237x2 videos of two mice (see Table \ref{tab:behaviour} for the description of social behaviour), which was used to study neurophysiological mechanisms involved in aggression and courtship in mice. It was recorded with synchronized top- and side-view cameras with the resolution of 640*480 pixels and the frame rate of 25Hz. Each video lasts about 10 minutes and was annotated frame by frame. 13 social behaviours was defined in this dataset, including 12 specific behaviours (shown in Table \ref{tab:behaviour}) and one otherwise unspecified behaviour (i.e., 'other'). In this paper, we use the public CRIM13 dataset with pose annotations (CRIM13-Skeleton) in \cite{nilsson2020simple}. It contains 64 top-view videos where each frame has 16 keypoints (each mouse has 8 keypoints), as shown in Fig.  \ref{fig:keypointimg2}(a) and \ref{fig:img_crim13}(a).  For each keypoint, it is represented by a tuple of $(X,Y,C)$, in which $(X,Y)$ is the 2D coordinates and $C$ denotes the confidence score. In our experiments, we only use 7 kypoints (0-6 in Fig. \ref{fig:keypointimg2}(a)) of each mouse due to low confidence on the tail end. Different from the original dataset \cite{burgos2012social}, CRIM13-Skeleton dataset is categorised into 12 behaviours where the behaviour 'human' is deleted. In our experiments, we randomly split the dataset into a training set of 51 videos and a test set of 13 videos.

\begin{table*}
\begin{center}
\caption{Ablation experiments for the Cross-Skeleton Node-level Interaction (CS-NLI) module on the CRIM13-Skeleton dataset. We present the classification accuracy ($\%$) of each behaviour,  average accuracy over all the behaviours, FLOPs and parameter number. The best performance is highlighted in bold.}
\label{tab:CS-NLI}
\begin{tabular}{p{2.2cm}p{0.8cm}p{0.4cm}p{1cm}p{0.4cm}p{0.4cm}p{0.4cm}p{0.4cm}p{0.4cm}p{0.4cm}p{0.4cm}p{0.4cm}p{0.4cm}p{0.7cm}p{0.7cm}p{0.7cm}}
\toprule
Methods\quad & Approach & Attack & Copulation & Chase & Circle & Drink & Eat & Clean &Sniff & Up & Walk away & Other &Average & Params &FLOPs\\
\midrule
Baseline  & 46.70 & 79.41  & 69.32 & 22.09 & 57.92 & 77.02 &53.02 & 77.17 & 69.68 & 71.87 & 47.59 & \textbf{73.79} & 62.13 & 1.05M  & 0.23G \\
\hline
\multicolumn{14}{c}{\textit{with dense geometric distance information }}\\
\hline
CS-NLI(w/o inter)  & 58.96	 &75.50	&71.65	&32.43	&56.29	&74.91 &60.72	&85.92	&64.54	&77.74	&49.41 &63.35 & 64.29 & 2.36M  & 0.33G \\
CS-NLI(w inter)  & 66.37	&71.17	&72.16	&\textbf{53.51}	&64.28	&\textbf{81.40}	&63.16	&76.90	&68.93	&83.47	&56.16	&53.33 & 67.57 & 2.37M  & 0.34G \\
\hline
\multicolumn{14}{c}{\textit{with dense velocity information }}\\
\hline
CS-NLI(w/o inter) & 56.60	&78.05	&71.53	&31.89	&55.97	&78.07	&55.09	&74.68	&67.80	&79.16	&50.47	&62.57 & 63.49  & 2.70M  & 0.35G\\
CS-NLI(w inter) &63.47	&76.39	&74.98	&34.19	&\textbf{67.99}	&79.65	&55.40	&82.27	&\textbf{70.39}	&74.54	&\textbf{66.46}	&41.52 & 65.60  & 2.71M  & 0.36G\\
\hline
\multicolumn{14}{c}{\textit{with dense geometric distance and velocity information }}\\
\hline
CS-NLI(w/o inter) & 62.04	&78.01	&75.55	&36.89	&59.75	&77.54	&\textbf{65.87}	 &80.55	&66.51	&80.71	&45.40	&56.67 & 65.46 & 2.89M  & 0.36G\\
CS-NLI(w inter) & \textbf{69.24}	&\textbf{81.81}	&\textbf{78.91}	&44.73	&66.23	&79.12	&57.70	&\textbf{86.87}	&65.72	&\textbf{85.12}	&59.10	&45.02 &\textbf{68.30} & 2.90M  & 0.37G\\
\bottomrule
\end{tabular}
\end{center}
\end{table*}

\subsubsection{PDMB-Skeleton Dataset}
\label{section4.1.2}
Our PDMB dataset was collected in collaboration with the biologists of Queen’s University Belfast of United Kingdom, for a study on motion recordings of mice with Parkinson’s disease (PD) \cite{jiang2021muti}. The neurotoxin 1-methyl-4-phenyl-1,2,3,6-tetrahydropyridine (MPTP) is used as a model of PD, which has become an invaluable aid to produce experimental parkinsonism since its discovery in 1983 \cite{jackson2007protocol}. By utilising MPTP-induced models, we aim to establish a direct link between changes in mouse social behaviours and the neurodegenerative processes associated with PD. The MPTP model allows us to mimic key aspects of PD pathology in mice, facilitating the study of behavioural changes that parallel the symptoms observed in human patients. Quantifying mouse social behaviours \cite{jiang2021muti} contributes to understanding how MPTP-induced neurodegeneration impacts specific behaviours. 
% This is not only crucial for advancing our comprehension of the disease but also for developing early diagnostic tools and intervention measures.

All experimental procedures were performed in accordance with the Guidance on the Operation of the Animals (Scientific Procedures) Act, 1986 (UK) and approved by the Queen’s University Belfast Animal Welfare and Ethical Review Body. This dataset consists of 12*3 annotated videos (6 videos for MPTP treated mice and 6 videos for control mice) recorded by using three synchronised Sony Action cameras (HDR-AS15) (one top-view and two side-view) with frame rate of 30 fps and 640*480 resolution. All videos contain 9 behaviours of two freely behaving mice and each video lasts around 10 minutes.

\begin{figure}[htbp]
\begin{center}
\includegraphics[width=8.2cm]{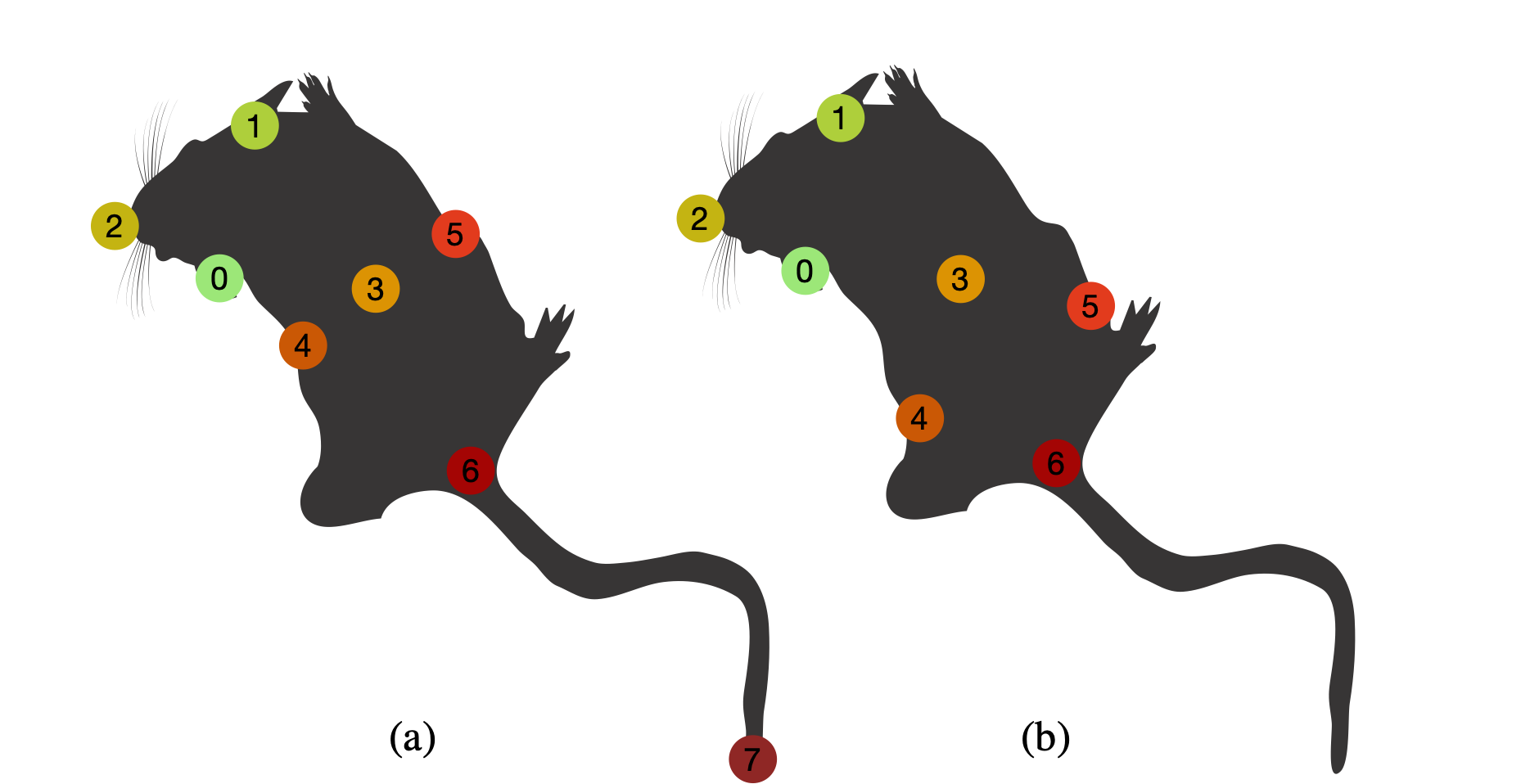}
\end{center}
\caption{Keypoint labels of (a) CRIM13-Skeleton and (b) PDMB-Skeleton datasets.
The CRIM13-Skeleton dataset contains 8 keypoints for each mouse (i.e., 0-left ear, 1-right ear, 2-snout, 3-centroid, 4-left lateral, 5-right lateral, 6-tail base and 7-tail end). The PDMB-Skeleton dataset contains 7 keypoints for each mouse (i.e., 0-left ear, 1-right ear, 2-snout, 3-centroid, 4-left hip, 5-right hip and 6-tail base).
}
\label{fig:keypointimg2}
\end{figure}

PDMB-Skeleton dataset is an extension of the original PDMB dataset \cite{jiang2021muti}, with added keypoint annotations, providing around 220,000 skeleton sequences. To obtain the location of each mouse keypoint, we also used the standard pose estimator, i.e., DeepLabCut \cite{Mathis2018}, to generate the locations and confidence scores of 7 defined keypoints on every frame of 12 top-view videos. We adopted the same training and testing dataset split scheme as in the PDMB dataset, evenly dividing the entire dataset into training and testing sets, resulting in 110,000 training samples.
Fig. \ref{fig:keypointimg2}(b) and \ref{fig:img_crim13}(b) show the keypoint locations in the PDMB-Skeleton dataset. More details about data annotation and dataset construction can be found in Supplementary B.

\vspace{-0.4cm} 

\subsubsection{Implementation Details}
\label{section4.1.3}
All the experiments are performed with PyTorch 1.4.0 on a server with an Intel Xeon CPU @ 2.40GHz and two 16GB Nvidia Tesla P100 GPUs. The parameters are optimised by the Adam algorithm. For the two datasets, we use the initial learning rate of 1e-4 and all the keypoint locations are normalised before training. No data augmentation is used for a fair performance comparison.
% Besides, we adopt the PyTorch sampler, i.e., Imbalanced Dataset Sampler\footnote{https://github.com/ufoym/imbalanced-dataset-sampler} to rebalance the class distributions for the two datasets during training.
$\beta$ and batch size $B_{b}$ are set to 0.5 and 128, respectively. As far as it concerns the model architecture, we use 3 cascaded CS-NLI+IAT modules, whose feature dimensions are 64, 128 and 256, respectively. The source code will be available at: \href{https://github.com/FeixiangZhou/CS-IGANet}{https://github.com/FeixiangZhou/CS-IGANet.}

\subsection{Ablation Study}
\label{section4.2}
In this section, we launch comprehensive experiments to investigate the effectiveness of each model component, i.e., Cross-Skeleton Node-level Interaction, Interaction-Aware Transformer, Self-supervision for cross-skeleton node similarity learning in our proposed CS-IGANet. We conduct ablation experiments on the CRIM13-Skeleton and PDMB-Skeleton datasets and use a 3-layer single-stream (keypoint) GCN-TCN network \cite{shi2019two} as our baseline model, where feature dimensions are 64, 128 and 256, respectively. Normally, classification accuracy is defined as the percentage of the samples that are correctly classified against the number of the overall samples. While it is a valid measure, this metric cannot disclose the characteristics of the datasets that have a severe imbalanced classification problem. Following \cite{jiang2021muti}, we here employ the averaging recognition rate per behaviour to better measure the system performance.

\begin{figure*}[htbp]
\begin{center}
\includegraphics[width=16.2cm]{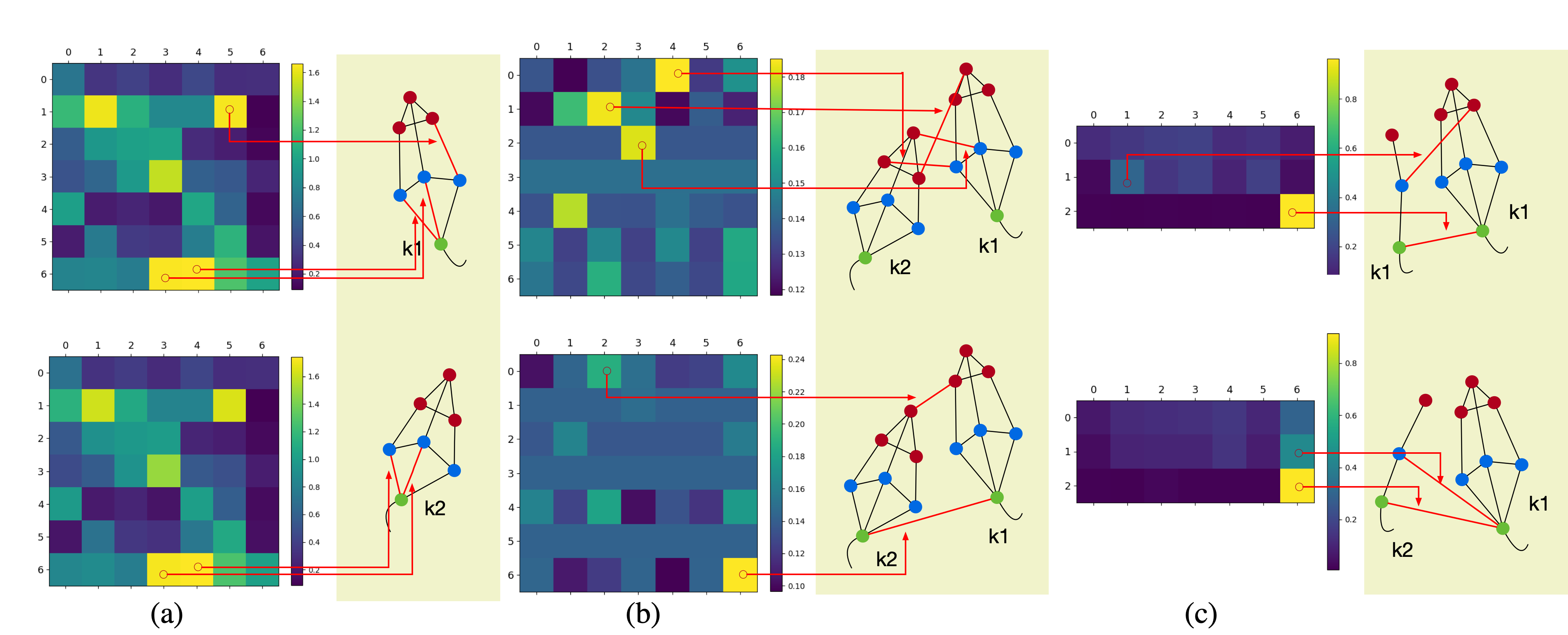}
\end{center}
 \caption{Visualisation of the learned topologies of a social behaviour sample 'approach' on the CRIM13-Skeleton dataset. (a) The topologies representing intra-skeleton interactions of mice $k_{1}$ (top) and $k_{2}$ (bottom). The number of keypoint $V$ is 7 and its configuration is shown in Fig. \ref{fig:img_crim13}. Here, we show the summation of the learned topologies on the three subsets generated by the partition strategy \cite{yan2018spatial}. (b) The topologies of bidirectional inter-skeleton interactions learned by our model, i.e., $\mathbf{A}_{k_{1}\to k_{2}}^{l=1}$ (top)  and $\mathbf{A}_{k_{2}\to k_{1}}^{l=1}$ (bottom). (c) The topologies of cross-skeleton interactions ($s_{1}$ to $s_{2}$) learned by our model, i.e., $\mathbf{A}_{s_{1}\to s_{2}}^{l=1}$ (top)  and $\widetilde{\mathbf{A}}_{s_{1}\to s_{2}}^{l=1}$ (bottom). For each category, we use red lines to highlight the interactions with high significance. Zoom in for the best visualisation.     }
\label{fig:matrix}
\end{figure*}

\begin{table*}[htbp]
\begin{center}
\caption{Ablation experiments for the Interaction-aware Transformer (IAT) on the CRIM13-Skeleton dataset.}
\label{tab:IAT}
\begin{tabular}{p{3cm}p{0.8cm}p{0.4cm}p{1cm}p{0.4cm}p{0.4cm}p{0.4cm}p{0.4cm}p{0.4cm}p{0.4cm}p{0.4cm}p{0.4cm}p{0.4cm}p{0.6cm}p{0.6cm}p{0.5cm} }
\toprule
Methods\quad & Approach & Attack & Copulation & Chase & Circle & Drink & Eat & Clean &Sniff & Up & Walk away & Other &Average  &Params &FLOPs\\
\midrule
Baseline  & 46.70 & 79.41  & 69.32 & 22.09 & 57.92 & 77.02 &53.02 & 77.17 & 69.68 & 71.87 & 47.59 & \textbf{73.79} & 62.13 & 1.05M  & 0.23G \\
IAT(I $\&$ w/o DC)  & 59.22 &78.07	&66.23	&42.84	&62.83	&80.53	&59.35	&88.07	&66.85	&82.92	&52.79	&47.60 &65.61 & 1.89M  & 0.35G \\
IAT(I $\&$ w DC)  & 62.53	&\textbf{80.37}	&70.80	&35.41	&65.97	&80.70	&64.50	&85.47	&64.48	&81.92	&\textbf{61.49}	&49.09 & 66.89 &1.97M  & 0.36G\\
IAT(I+M $\&$ w DC)   & \textbf{65.53}	&78.46	&71.43	&51.62	&57.99	&81.05	&56.74	&\textbf{88.46}	&67.10	&81.94	&60.13	&44.30 & 67.06 &1.97M  & 0.36G\\
IAT(I+A $\&$ w DC)    & 63.80	&74.20	&\textbf{77.02}	&50.95	&63.27	&\textbf{82.46}	&60.45	&83.43	&65.53	&83.72	&60.23	&48.21 & 67.77 &1.97M  & 0.36G\\
IAT(I+A+M $\&$ w DC)  & 61.96	&77.05	&72.98	&52.84	&62.52	&81.58	&59.90	&81.99	&\textbf{71.63}	&81.34	&60.60	&49.92 & 67.85 &1.97M  & 0.36G\\
IAT($G$(I,A,M) $\&$ w DC)   & 63.20	&75.90	&73.83	&\textbf{55.0}	&\textbf{75.91}	&79.65	&\textbf{67.29}	&83.95	&67.21	&\textbf{84.46}	&60.60	&41.03 &\textbf{ 69.0} &2.14M  & 0.39G\\
\bottomrule
\end{tabular}
\end{center}
\end{table*}

\begin{figure*}[htbp]
\begin{center}
\begin{tabular}{c}
\includegraphics[width=17cm]{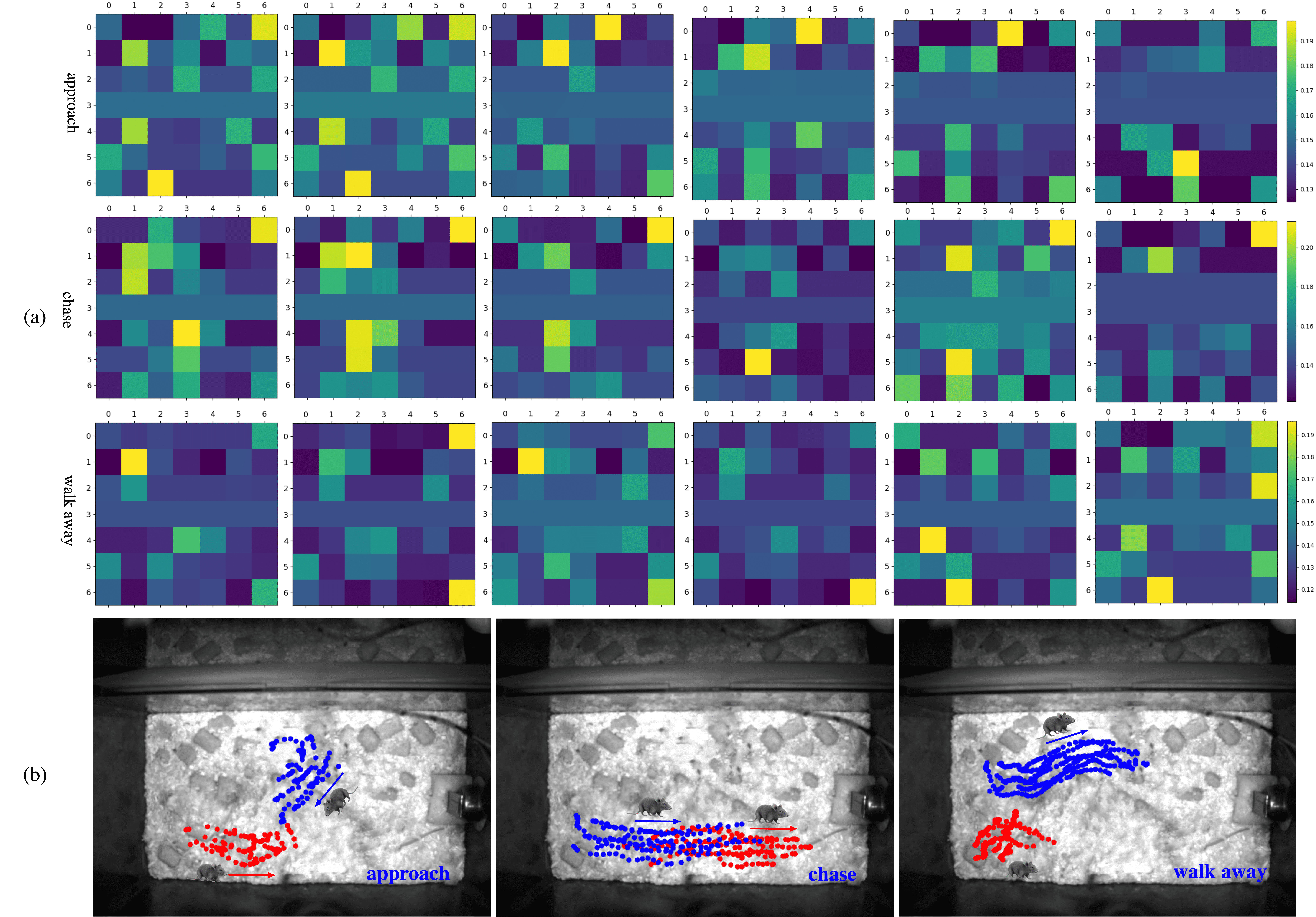}
% \hspace{-8mm}
\end{tabular}
\end{center}
\caption{Learned topologies (i.e., inter-skeleton interaction $\mathbf{A}_{k_{1}\to k_{2}}^{l=1}$ ) by our CS-NLI module for three sample social behaviours (e.g., approach, chase and walk away) (a) and the motion trajectories for three behaviours in the CRIM13-Skeleton dataset (b). For each behaviour shown in (a), each topology refers to the social interaction between mice in the current frame. The corresponding motion trajectories of keypoints are shown in (b). In (b), blue and red points indicate the keypoints of the resident mouse and the intruder, respectively. Blue and red arrows represent the direction of motion. More examples of motion trajectories can be found in Fig. \ref{fig:trajectory_eachbehaviour}. Best viewed in colour. }
\label{fig:trajectory}
\end{figure*}

\subsubsection{Effects of Cross-Skeleton Node-level Interaction}
\label{section4.2.1}
Here, we investigate the influences of the proposed Cross-Skeleton Node-level Interaction module. We study the effects of dense geometric distance in Eq. (\ref{equation:bone}), dense velocity in Eq. (\ref{equation:velicity}) as well as inter-skeleton interaction (denoted as inter-skeleton) block, presented in Table \ref{tab:CS-NLI}. We observe that although the baseline method only using the dense skeleton possesses the fewest parameters and FLOPs, it exhibits the poorest performance, especially for behaviours such as 'approach', 'chase' and 'walk away'. The low accuracy is due to the fact it only focuses on intra-skeleton interaction without considering the social interaction between mice. In addition, directly modelling cross-skeleton interaction of the same mouse based on dense geometric distance or velocity information, without inter-skeleton interaction, can help slightly improve the average accuracy. The performance can be further improved to 65.46$\%$ by fusing the two types of information while the number of parameters only increases by 0.19M and the time complexity increases by 0.01G, confirming that the dense geometric distance and velocity information are beneficial to cross-skeleton interaction encoding. With respect to the inter-skeleton interaction modelling, we witness that the models with inter-skeleton interaction based on different types of information all achieve better performance than the models without such interaction. It is noticeable that the addition of inter-skeleton interaction almost negligibly increases the model's complexity in terms of parameter count and FLOPs but significantly improves the average accuracy. In particular, for social behaviours such as 'approach', 'chase' and 'walk away', we can obtain 7.2$\%$, 7.84$\%$ and 13.7$\%$ improvements, respectively by modelling inter-skeleton interaction (see the last two rows). Such significant improvements are due to the strong interaction modelling ability of our CS-NLI module. By jointly modelling intra-, inter- and cross-skeleton interactions based on multi-order dense information, our method achieves the best average accuracy of 68.3$\%$, which is a 6.17$\%$ improvement against the baseline model.

% We observe that the baseline method
% achieves relatively low accuracy for behaviours such as 'approach', 'chase' and 'walk away'. The low accuracy is due to the fact it only focuses on intra-skeleton interaction without considering the social interaction between mice. In addition, directly modelling cross-skeleton interaction of the same mouse based on dense geometric distance or velocity information, without inter-skeleton interaction and cross-skeleton interaction of different mice, can help slightly improve the average accuracy. The performance can be further improved to 65.46$\%$ by fusing the two types of information, confirming that the dense geometric distance and velocity information are beneficial to cross-skeleton interaction encoding. With respect to the inter-skeleton interaction modelling, we witness that the models with inter-skeleton interaction based on different types of information achieve better performance than the models without such interaction. In particular, for social behaviours such as 'approach', 'chase' and 'walk away', we can obtain 7.2$\%$, 7.84$\%$ and 13.7$\%$ improvements, respectively by modelling inter-skeleton interaction (see the last two rows). Such significant improvements are due to the social interaction encoded by our CS-NLI module effectively. By jointly modelling intra-, inter- and cross-skeleton interactions based on multi-order dense information, our method achieves the best average accuracy of 68.3$\%$, which is a 6.17$\%$ improvement against the baseline model.

\begin{table*}
\begin{center}
\caption{Ablation experiments for the self-supervision of Cross-Skeleton Node Similarity Learning on the CRIM13-Skeleton dataset.}
\label{tab:self}
\begin{tabular}{p{3.2cm}p{0.8cm}p{0.5cm}p{1cm}p{0.5cm}p{0.5cm}p{0.5cm}p{0.5cm}p{0.5cm}p{0.5cm}p{0.5cm}p{0.5cm}p{0.5cm}p{1cm}}
\toprule
Methods\quad & Approach & Attack & Copulation & Chase & Circle & Drink & Eat & Clean &Sniff & Up & Walk away & Other &Average\\
\midrule
CS-NLI(w/o self, $\lambda=0$)  & \textbf{69.24}	&\textbf{81.81}	&\textbf{78.91}	&44.73	&66.23	&79.12	&57.70	&\textbf{86.87}	&65.72	&85.12	&59.10	&45.02 &68.30\\
CS-NLI(w self, $\lambda=0.1$)  & 59.65	&79.62	&74.83	&57.16	&64.91	&80.87	&64.19	&82.86	&63.45	&83.07	&\textbf{75.32}	&49.79 & 69.65\\
CS-NLI(w self, $\lambda=0.5$)  & 62.40	&77.69	&74.55	&\textbf{61.62}	&\textbf{79.43}	&\textbf{83.33}	&\textbf{65.60}	&85.17	&63.40	&87.07	&63.94	&42.76 & \textbf{70.58}\\
CS-NLI(w self, $\lambda=1$)  &67.50	 &81.43	&65.73 &48.66	 &77.04	& 82.81 &64.09	 &84.96 &\textbf{66.72}	& \textbf{89.04} &59.44	 &45.25 & 69.39\\
CS-NLI(w self, $\lambda=1.5$)  &59.68	& 76.45	& 80.03 &47.43	& 77.61	& 83.33 &55.15 &80.52	& 61.75 &86.90	& 66.24	& \textbf{52.51} & 68.97\\
\bottomrule
\end{tabular}
\end{center}
\end{table*}

We also visualise the relevant interaction patterns identified by our CS-NLI module, including the intra-, inter-, and cross-skeleton interactions. Fig. \ref{fig:matrix} shows the corresponding topologies of a behaviour sample 'approach' in the top branch of Fig. \ref{fig:CS-NLI} (i.e., skeleton $s_{1}$ to skeleton $s_{2}$). The values close to 0 indicate weak relationships between keypoints and vice versa. From Fig. \ref{fig:matrix}(a), we observe that the two topologies representing the intra-skeleton interactions of mice are very similar, where the correlations between some keypoint pairs are relatively strong, e.g. the correlation between the centroid and the tail base, and the correlation between the left lateral and the tail base. However, these independent relationships derived from each mouse are insufficient to be exploited to encode complex social interactions. The inter-skeleton interaction modelling is able to learn new interactions between mice that the independent skeleton graph cannot provide, as shown in Fig. \ref{fig:matrix}(b). For instance, our CS-NLI module pays much attention to the interactions between the tail bases of mice, and between the left ear and the snout when considering the effect of mice $k_{2}$ on $k_{1}$. Moreover, in Fig. \ref{fig:matrix}(c), our CS-NLI module further models the cross-skeleton interactions, where the tail bases of the same mice or different mice from different scales are highly related. To examine the difference of the topologies during training, we also show topologies learned by our CS-NLI module that is not fully trained, as shown in Fig. \ref{fig:matrix_inter}. We observe that the model generates a relatively dense fully connected graph at the beginning of training, especially for the inter- and cross-interactions, where interactions may not be related to behaviours.
On the contrary, our final model tends to better focus on behaviour related interactions. To show how our CS-NLI module works, we display the learned topologies representing the inter-skeleton interaction, i.e., $\mathbf{A}_{k_{1}\to k_{2}}^{l=1}$, in Fig. \ref{fig:trajectory}. From this figure, we observe that the CS-NLI module gives much attention to the interactions between mice, e.g., left lateral and left ear for 'approach', tail base and left ear for 'chase', and tail bases for 'walk away'.

\subsubsection{Effects of Interaction-aware Transformer}
\label{section4.2.2}

In order to validate the effectiveness of our proposed IAT module, we design six structures using the baseline model. IAT (I $\&$ w/o DC) refers to the case where we only keep the encoder of the IAT and use the graph-level representation aggregated by $IAT(\cdot)$ ($I$) to perform behavioural classification, while IAT (I $\&$ w DC) refers to a structure with the encoder and the decoder. IAT (I+M $\&$ w DC) and IAT (I+A $\&$ w DC) mean that we enhance the graph-level representation by combining graph-level representation from the encoder and that generated by spatial max pooling and average pooling, respectively, where we simply use the sum operation to fuse different representations. The last IAT ($G$(I,A,M) $\&$ w DC) refers to the structure that models the interactions between multi-level graph representations. From Table \ref{tab:IAT}, we observe that the IAT without any decoder achieves higher accuracy than the baseline model for all 8 behaviours, indicating that the intra-skeleton interaction of each mouse and inter-skeleton interactions between mice are important to graph-level representation learning. The performance is further improved by constructing an encoder-decoder framework, resulting in the highest accuracy of 80.37$\%$ and 61.49$\%$ for 'attack' and 'walk away', respectively. This is because the node-level representation can be adaptively updated by our proposed dual-path decoder, before it is fed into the next layer, which helps to identify higher-level features. In addition, three models combining different graph-level representations through a straightforward summation operation all outperform IAT (I $\&$ w DC), without incurring an increase in the number of parameters and FLOPs, suggesting that different graph-level representations carry complementary spatio-temporal information that helps improve the identification. Instead of fusing different graph-level representations by the sum operation, we explore the structural relations by our proposed interaction-aware self-attention unit, leading to a 1.15$\%$ improvement against IAT (I+A+M $\&$ w DC). Regarding our network involving two skeletons, we fuse different graph-level representations from two skeleton branches by the interaction-aware self-attention module. In contrast to the baseline, our proposed IAT 
demonstrates a noteworthy 6.87$\%$ improvement in average accuracy, although the computational complexity of this approach is more than twice that of the GAP-based method.

\subsubsection{Effects of Self-supervision for Cross-Skeleton Node Similarity Learning}
\label{section4.2.3}
In this subsection, we study the effect of the proposed auxiliary self-supervised loss function, controlled by the hyper-parameter $\lambda$. To analyse the impact of this parameter, we train several models (i.e., CS-NLI module) with different values of $\lambda$. As shown in Table \ref{tab:self}, all models with self-supervision ($\lambda =0.1,0.5,1,1.5$) lead to an improvement over the baseline without self-supervision. Increasing $\lambda$ from 0 to 0.5 significantly improves the accuracy. This is mainly because the important attributes (i.e., similarity) underlying cross-skeleton pairwise nodes are explicitly exploited in the node representation learning.
When $\lambda =0.5$, we achieve significant improvements on the accuracy of 'chase', 'circle', 'drink' and 'eat', and the highest average accuracy of 70.58$\%$. However, there is significant degradation in the performance when we increase $\lambda$ to 1.5. This drop is due to the fact that the self-supervised loss severely penalises the inherent attributes of node pairs from different skeletons. Hence, our default value is $\lambda =0.5$.

\begin{table}[htbp]
\begin{center}
\caption{Ablation experiments for the CS-NLI, IAT and self-supervision modules on the CRIM13-Skeleton and PDMB-Skelton datasets.}
\label{tab:ablationstudy_all}
\begin{tabular}{p{2cm}p{0.4cm}p{0.4cm}p{1cm}p{0.6cm}p{0.5cm}p{0.6cm}}
\toprule
Dataset &CS-NLI &IAT & Self-supervision & Average &Params &FLOPs\\
\midrule

\multirow{5}*{CRIM13-Skeleton}&  &  & & 62.13 & 1.05M  & 0.23G\\
&$\surd$  &   & & 68.30 &2.90M &0.37G\\
&  & $\surd$  &  & 69.0 & 2.14M &0.39G\\
&$\surd$  &   & $\surd$& 70.58 &2.90M &0.37G\\
&$\surd$  & $\surd$ & $\surd$ & \textbf{73.75}&4.91M &0.61G\\
\midrule
\multirow{5}*{PDMB-Skeleton}&  &  & & 52.67  & 1.05M  & 0.23G\\
&$\surd$  &   & & 59.24 &2.90M &0.37G\\
&  & $\surd$  &  & 60.03 & 2.14M &0.39G\\
&$\surd$  &   & $\surd$& 60.59&2.90M &0.37G\\
&$\surd$  & $\surd$ & $\surd$ & \textbf{62.33}&4.91M &0.61G\\
\bottomrule
\end{tabular}
\end{center}
\end{table}

\begin{figure}[htbp]
\begin{center}
\includegraphics[width=8cm]{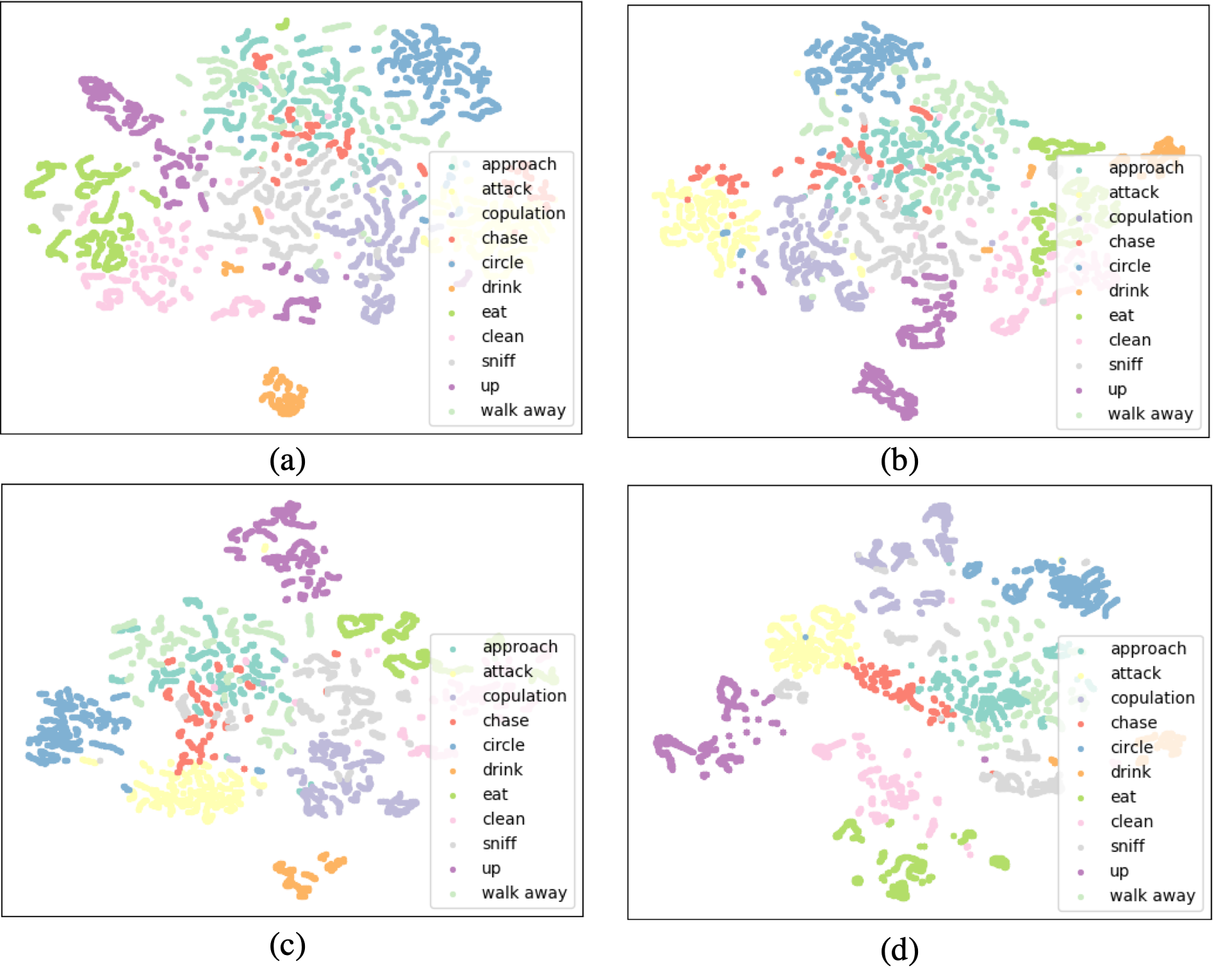}
\end{center}
\caption{t-SNE visualization of the features learned by (a) 2s-AGCN, (b) MS-G3D, (c) MV-IGNet, (d) Our method. Each point represents a skeleton sequence. We show 11 behaviour classes of the CRIM13-Skeleton dataset, indicated by colours.}
\label{fig:featuretsne}
\end{figure}

We also investigate the contribution of the proposed CS-NLI, IAT and Self-supervision modules to the whole network on both datasets. In addition to adding each proposed module to the baseline model separately, we further employ the proposed modules applied to the baseline model simultaneously. The results are reported in Table \ref{tab:ablationstudy_all}. On the two datasets, our method achieves the highest average accuracy, 73.75$\%$ and 62.33$\%$ respectively, with the three proposed modules applied simultaneously, which are of 11.62$\%$ and 9.66$\%$ increments compared to the baseline model. In terms of the model complexity  (FLOPs and number of parameters),  applying either CS-NLI or IAT to the baseline results in an approximately twofold increase. However, the incorporation of Self-supervision does not lead to an additional computational cost. Though the model complexity of our method with the three components is higher than the baseline, it receives a much higher accuracy.

\begin{table*}[htbp]
\begin{center}
\caption{Comparisons with state-of-the-art methods on the CRIM13-Skeleton dataset in classification accuracy ($\%$), FLOPs and and parameter number. The best performance is highlighted in bold.}
\label{tab:comparision_crim13}
\begin{tabular}{p{2.7cm}p{0.8cm}p{0.5cm}p{0.9cm}p{0.5cm}p{0.5cm}p{0.5cm}p{0.5cm}p{0.5cm}p{0.5cm}p{0.5cm}p{0.5cm}p{0.5cm}p{0.6cm}p{0.5cm}p{0.6cm}}
\toprule

Methods\quad & Approach & Attack & Copulation & Chase & Circle & Drink & Eat & Clean &Sniff & Up & Walk away & Other &Average   &Params &FLOPs\\
\midrule
\midrule
ST-GCN\cite{yan2018spatial}   & 34.34	&75.68	&68.97	&35.56	&34.65	&73.33	&45.87	&73.53	&64.31	&69.75	&26.27	&\textbf{75.80} & 56.51 &3.07M &0.61G\\
2s-AGCN\cite{shi2019two} 	&51.03	&83.40	&75.84	&36.46	&53.40	&75.61	&63.65	&76.71	&75.80	&76.25	&40.04 &51.15 & 63.28 &6.87M &1.39G\\
SGN\cite{zhang2020semantics} & 49.57	&80.43	&74.90	&34.90	&66.42	&75.44	&63.26	&83.83	&66.19	&75.67	&41.78	&57.41 & 64.15 &0.66M &0.20G\\
PA-ResGCN\cite{song2020stronger}   & 57.04	&74.56	&77.81	&22.98	&51.64	&80.88	&55.90	&82.99	&75.42	&85.15	&40.44	&69.16 &64.50 &3.46M &0.70G\\
MS-G3D\cite{liu2020disentangling} & 57.38	&80.08	&69.79	&56.61	&75.28	&65.79	&\textbf{72.74}	&66.00	&55.96	&82.00	&52.03	&49.29 & 65.24  &5.54M &1.83G\\
MV-IGNet\cite{wang2020learning}  & 55.33	&73.77	&\textbf{77.81}	&48.78	&60.31	&81.75	&51.59	&80.81	&\textbf{78.43}	&83.40	&57.83	&64.56 & 67.86 &1.80M &0.54G\\

ST-TR\cite{plizzari2021skeleton}  & 45.92	&74.86	&74.93	&44.93	&78.43	&79.12	&54.49 &84.84	&72.85	&\textbf{89.19}	&60.69	&23.04 &65.27 &12.0M &2.40G\\

EfficientGCN\cite{song2022constructing} &52.99	&78.47	&75.12	&33.12	&44.97	&81.23	&55.36	&80.20	&77.96	&83.23	&55.29	&66.39 &65.36 &2.02M &0.62G\\

CTR-GCN\cite{chen2021channel}  & 52.33	&79.45	&75.89	&41.59	&53.08	&66.67	&55.50	&68.12	&73.70	&78.71	&48.83	&60.58 & 62.83 &5.66M &1.22G\\
InfoGCN\cite{chi2022infogcn}  & 51.98	&83.45	&77.21	&54.04	&62.14	&76.32	&53.07	&70.82	&57.76	&74.59	&49.49	&64.01 & 64.57 &6.08M &1.17G\\
STEP CATFormer\cite{huu2023step} & 50.49	&75.52	&70.00	&51.09	&61.03	&71.43	&60.45	&64.87	&68.75	&73.54	&44.21	&55.26 & 62.22 &39.8M &1.76G\\
2s-DRAGCN \cite{zhu2021dyadic} & 53.03 & 75.52 & 75.90 & 54.41 & 55.72 & 74.21 & 51.47 & 75.44 & 61.76 & 70.02 & 53.20 & 60.56 & 63.44 & 7.41M & 1.54G \\
2P-GCN \cite{li2022two} & 57.35 & 78.99 & 70.04 & 53.79 & 65.47 & 72.11 & 56.86 & 69.49 & 65.05 & 77.53 & 54.72 & 58.63 &65.00 & 1.45M & 0.64G \\
ISTA-Net\cite{wen2023interactive} & 58.34	&78.57	&71.08	&57.59	&64.78	&70.22	&56.50	&72.56	&68.54	&75.51	&57.65	&56.87 & 65.68 &5.68M &3.18G\\

Ours(CS-NLI+self)   & 62.40	&77.69	&74.55	&61.62	&79.43	&\textbf{83.33}	&65.60	&85.17	&63.40	&87.07	&63.94	&42.76 & 70.50  &2.90M &0.37G\\
Ours(IAT)   & 63.20	&75.90	&73.83	&55.0	&75.91	&79.65	&67.29	&83.95	&67.21	&84.46	&60.60	&41.03 & 69.0  &2.14M &0.39G\\
% Ours(CS-IGANet) & \textbf{64.81}	&\textbf{83.49}	&76.06	&\textbf{62.70}	&\textbf{80.88}	&79.30	&65.91	&\textbf{86.25}	&73.23	&85.44	&\textbf{69.46}	&49.11 & \textbf{73.05}\\
Ours(CS-IGANet) & \textbf{67.30}	&\textbf{84.30}	&75.30	&\textbf{69.32}	&\textbf{82.08}	&81.75	&63.44	&\textbf{88.20}	&73.47	&87.69	&\textbf{66.79}	&45.38 & \textbf{73.75}  &4.91M &0.61G\\
\bottomrule
\end{tabular}
\end{center}
\end{table*}

\begin{table*}[htbp]
\begin{center}
\caption{Comparisons with state-of-the-art methods on the PDMB-Skeleton dataset in classification accuracy ($\%$), FLOPs and and parameter number. The best performance is highlighted in bold.}
\label{tab:comparision_pdmb}
\begin{tabular}{p{2.7cm}p{1cm}p{0.6cm}p{0.6cm}p{0.6cm}p{0.6cm}p{0.6cm}p{0.6cm}p{0.6cm}p{0.6cm}p{1cm}p{0.6cm}p{0.6cm}}
\toprule

Methods\quad & Approach & Chase & Circle & Eat & Clean &Sniff & Up & Walk away & Other &Average  &Params & FLOPs\\
\midrule
\midrule
ST-GCN\cite{yan2018spatial}   & 52.72	&44.24	&50.62	&53.24	&45.84	&30.61	&66.74	&38.09	& \textbf{81.69} &51.53 &3.07M &0.61G\\

2s-AGCN\cite{shi2019two}  & 54.84	&42.17	&53.89	&51.08	&48.36	&50.40	&74.04	&41.78	&61.95 &53.17 &6.87M &1.39G\\
SGN\cite{zhang2020semantics}  & 53.96	&41.47	&47.19	&56.56	&46.77	&44.54	&73.13	&40.82	&76.13 &53.40  &0.66M &0.20G\\
PA-ResGCN\cite{song2020stronger}  & 59.06	& 45.16	& 52.34	& 58.99	& 50.98	& 43.11	& 67.71	& 50.35	&  78.84 & 56.28 &3.46M &0.70G\\
MS-G3D\cite{liu2020disentangling}  & 58.10	&44.01	&61.99. &57.55	&47.10	&70.0	&69.13	&37.36	&48.92 & 54.91  &5.54M &1.83G\\
MV-IGNet\cite{wang2020learning}  & 51.47	&45.39	&51.40	&55.40	&57.22	&42.28	&\textbf{89.45}	&39.15	&75.17 &56.33  &1.80M &0.54G\\
ST-TR\cite{plizzari2021skeleton}  & 56.33	&50.69	&51.87	&\textbf{59.71}	&46.81	&41.49	&72.76	&49.95	&61.25 & 54.54  &12.0M &2.40G\\
EfficientGCN\cite{song2022constructing}  & 56.28	&40.09	&51.56	&51.80	&51.28	&44.77	&74.18	&37.91	&80.25 & 54.23 &2.02M &0.62G\\
CTR-GCN\cite{chen2021channel} & 46.48	&48.39	&50.93	&54.68	&50.51	&68.83	&62.69	&38.52	&50.55 & 52.32  &5.66M &1.22G\\
InfoGCN\cite{chi2022infogcn} & 51.47	&45.39	&53.40	&58.40	&57.22	&52.28	&74.54	&39.15	&51.69 & 53.73  &6.08M &1.17G\\
STEP CATFormer\cite{huu2023step} & 50.98	&41.71	&52.84	&61.15	&54.99	&52.24	&60.38	&42.30	&60.51 & 53.01  &39.8M &1.76G\\
2s-DRAGCN \cite{zhu2021dyadic} & 55.47 & 50.39 & 53.40 & 51.40 & 52.22 & 50.23 & 70.45 & 42.15 & 59.17 & 53.88 &7.41M & 1.54G\\
2P-GCN \cite{li2022two} & 57.43 & 49.56 & 57.16 & 54.57 & 46.23 & 54.54 & 65.36 & 48.82 & 56.37 & 54.45 & 1.45M & 0.64G \\
ISTA-Net\cite{wen2023interactive} & 59.89	&54.37	&58.94	&55.08	&53.04	&56.51	&62.29	&50.47	&52.01 & 55.84  &5.68M &3.18G\\
Ours(CS-NLI+self) & 65.83	&60.83	&62.93	&56.12	&54.03	&63.60	&75.16	&52.44	&54.40 & 60.59  &2.90M &0.37G\\
Ours(IAT) & 61.34	&60.37	&56.70	&58.99	&53.04	&68.22	&74.52	&50.68	&56.25 & 60.03  &2.14M &0.39G\\
Ours(CS-IGANet)  &\textbf{66.43}	&\textbf{63.13} &\textbf{63.40}	&57.55	&\textbf{58.08}	&\textbf{71.33}	&70.53	&\textbf{53.66}	&56.88 & \textbf{62.33}  &4.91M &0.61G\\
\bottomrule
\end{tabular}
\end{center}
\end{table*}

\subsection{Comparison to State-of-the-Art Approaches}
\label{section4.3}
In this section, we compare the proposed model against several state-of-the-art graph-based methods and transformer-based methods on two datasets: CRIM13-Skeleton and PDMB-Skeleton. In our experiments, we follow the default settings of all existing methods to ensure a fair comparison. We use the same configuration parameters, multi-modality information and hyperparameters as specified by the authors of those methods. Specifically, the selected methods include single-stream \cite{yan2018spatial,wen2023interactive} and multi-stream \cite{shi2019two,zhang2020semantics,song2020stronger,liu2020disentangling,wang2020learning,plizzari2021skeleton,chen2021channel,song2022constructing,chi2022infogcn,huu2023step} frameworks. Most methods (e.g., 2s-AGCN (2-stream, denoted as 2s), MS-G3D(2s), MV-IGNet(2s), CTR-GCN(4s), InfoGCN(4s), STEP CATFormer(4s)) employ a multi-stream fusion framework, where different modalities  (e.g., joint/keypoint, bone, joint motion/velocity, and bone motion) are used as inputs to separately train the same network to obtain better results. Although these networks have proven effective, multiple separate networks will increase the number of parameters. Some methods (e.g., SGN(2s), EfficientGCV(3s), ST-TR(2s), PA-ResGCN(3s)) fuse several types of information in the early stage (in input) to reduce the computational cost caused by the multi-stream structure. We also compare three methods of interactive action recognition, i.e., 2s-DRAGCN \cite{zhu2021dyadic}, 2P-GCN \cite{li2022two} and  ISTA-Net \cite{wen2023interactive}. Similar to ST-GCN \cite{yan2018spatial} and ISTA-Net \cite{wen2023interactive}, we only employ the keypoint information as input without explicitly fusing other information. Notably, our proposed method can be considered as the multi-stream structure based on early fusion but we extract dense geometric distance and velocity information to better describe nuanced social interactions between mice, which is different from all the above approaches. The results on two datasets are presented in Tables \ref{tab:comparision_crim13} and \ref{tab:comparision_pdmb}, respectively.

% In particular, a wide range of technologies were proposed to improve the ability of skeleton encoding in the selected skeleton-based action recognition approaches. Specifically, In ST-GCN \cite{yan2018spatial}, the presence of edges denoting the relationships between keypoints is pre-defined by human based on prior knowledge, and all the edges keep fixed during training. 2s-AGCN \cite{shi2019two} is a two-stream structure that focuses on learning a content adaptive graph to aggregate node information. Additionally, MS-G3D \cite{liu2020disentangling} explores complex spatial-temporal dependency by the spatial-temporal graph convolutional operator. CTR-GCN \cite{chen2021channel} aims to obtain channel-wise topologies through learning a shared topology (parameterized adjacency matrix that serves as topological priors) and channel-specific correlations simultaneously. We also compare a multi-view model named MV-IGNet \cite{wang2020learning} using two different skeleton topologies as input. In our experiments, we utilize the defined dense and sparse mouse skeletons as the input. Another popular model called ST-TR \cite{plizzari2021skeleton} is based on transformer self-attention mechanism to extract information from keypoint motion patterns and their correlations. Finally, we also choose three lightweight models, i.e., SGN \cite{zhang2020semantics}, PA-ResGCN \cite{song2020stronger} and EfficientGCN \cite{song2021constructing} for comparisons. For the all above existing methods, we follow their default settings in our experiments.

As shown in Table \ref{tab:comparision_crim13}, Our proposed modules, along with their combined architecture, demonstrate superior performance compared to other state-of-the-art models in terms of average accuracy while having relatively few parameters and FLOPs. This is because our method jointly models the intra-, inter- and cross-skeleton interactions, and dynamically learns graph-level representation of mouse social behaviours, which is very effective in representation learning of mouse social behaviour. Although ST-GCN \cite{yan2018spatial} achieves the highest classification accuracy on 'other', its average accuracy of all the behaviours is the lowest among these methods. This is because it only uses a fixed skeleton topology to model the relations between keypoints of each mouse, limiting its ability to encode/decode intra-skeleton interaction for some specific behaviours, such as 'eat' and 'up'. Compared with CTR-GCN \cite{chen2021channel} using mul-stream fusion, our CS-IGANet exhibits a substantial performance advantage, where the average accuracy holds 10.92$\%$ improvement and the computational complexity, measured in FLOPs, is reduced from 1.22G to 0.61G. This suggests that dynamic refinement of channel-wise topology is not powerful enough to maintain the quality of mouse social behaviour representation. We also notice that, among the 12 existing methods, MV-IGNet \cite{wang2020learning} achieves the best performance with relatively fewer parameters and lower computational costs, but there is still a large gap (i.e., 5.89$\%$) between MV-IGNet and our CS-IGANet. Compared to Transformer-based methods such as ST-TR \cite{plizzari2021skeleton} and STEP CATFormer \cite{huu2023step}, our method consistently exhibits superior performance and the parameter numbers and computational costs of them are several times that of our method. Another Transformer-based method, i.e., ISTA-Net \cite{wen2023interactive}, models interactive relations of diverse interacting subjects. Although it improves the accuracy of certain behaviours such as 'chase', the overall performance still lags behind ours by approximately 8$\%$ and the model complexity is higher than that of our method. In addition, for social interactions such as approach, chase and walk away, our proposed CS-IGANet improves the accuracy with large margins of 8.96$\%$, 11.73$\%$ and 6.1$\%$, respectively, compared with their close competitors. We also show the confusion matrix of our CS-IGANet on the CRIM13-Skeleton dataset, as shown in Fig. \ref{fig:confusionm}(a).

Notably, the accuracy for behaviour 'other' is significantly lower compared to almost all other methods. The ‘other’ category refers to instances where no behaviour of interest is occurring and it normally constitutes over 50\% of the entire dataset \cite{burgos2012social}. However, in mouse behaviour research, accurately identifying other meaningful behaviours or interactions is more crucial \cite{jiang2021muti,burgos2012social}. Our method improves the accuracy of most meaningful behaviours by effectively capturing the spatio-temporal dynamics of social interactions. On the other hand, methods, such as ST-GCN, lack the ability to model meaningful social interactions. Consequently, these methods tend to classify various behaviours as the ‘other’ category, leading to higher accuracy in this category but poor performance in meaningful behaviour classes. Despite the lower accuracy in the ‘other’ category, our proposed CS-IGANet significantly improves the average accuracy across all behaviour classes. We believe this trade-off highlights the effectiveness of our approach in achieving a more balanced and accurate overall classification, particularly in the more accurate recognition of meaningful behaviours.

Additionally, we present in Fig. \ref{fig:featuretsne} the t-SNE visualisation of the representations learned by our model and other 3 state-of-the-art methods (i.e., 2s-AGCN, MS-G3D and MV-IGNet). For our model, the representation is the concatenation of graph-level outputs of different layers, i.e., $\widetilde{\mathbf X}_{g}^{\text{cross}}$. Our proposed CS-IGANet leads to better separation of the 11 behaviour classes. In particular, for some similar behaviours such as 'approach' and 'walk away', our model can better distinguish them.

As for the PDMB-Skeleton dataset, our approach also achieves the state-of-the-art performance with average accuracy of 62.33$\%$, which is a 6$\%$ improvement compared with the closest competitor, i.e., MV-IGNet. In addition, our CS-IGANet significantly outperforms the other state-of-the-art methods on behaviours 'approach', 'chase', 'circle' and 'walk away', and achieves comparable performance on 'clean' and 'sniff', compared to \cite{wang2020learning} and \cite{liu2020disentangling}. The confusion matrix of our CS-IGANet on the PDMB-Skeleton dataset is shown in Fig. \ref{fig:confusionm}(b). 
To further evaluate the generalisation capability of our proposed method, we also conduct experiments on two human datasets (NTU-Interaction \cite{shahroudy2016ntu} and NTU120-Interaction \cite{liu2019ntu}), as shown in Tab. \ref{tab:comparison_human}. Our method continues to exhibit comparable or competitive performance. 

\section{Conclusion}
In this work, we have presented a novel architecture called Cross-Skeleton Interaction Graph Aggregation Network (CS-IGANet) for representation Learning of mouse social behaviour. Cross-Skeleton Node-level Interaction module (CS-NLI) strengthens the node-level representation of each mouse by modelling intra-, inter- and cross-skeleton interactions in a unified way. We also designed a novel Interaction-Aware Transformer (IAT) to hierarchically aggregate node-level representation into graph-level representation of social behaviour, and adaptively update the node-level representation, which is guided by our interaction-aware self-attention unit. An auxiliary self-supervised learning task was also proposed to focus on the similarity between cross-skeleton pairwise nodes, enhancing the representation ability of our model. Experimental results on CRIM13-Skeleton and PDMB-Skeleton datasets demonstrated that the proposed approach outperformed most of the baseline methods. Our proposed solution is currently working on two mice cases but is extendable to three or more mice. The only difference is scaling up the complexity of computation. In addition, the proposed method for modelling social interactions of mice can be potentially extended to collaborative human behaviour prediction, especially for some scenarios involving nuanced behaviour patterns. For example, in collaborative work environments, understanding subtle cues and interactions among individuals is crucial for predicting actions and ensuring effective collaboration. Our future work in this direction could involve refining our approach to capture human-specific social dynamics and evaluating its performance in collaborative scenarios. We also plan to improve the efficiency and scalability of our approach to better meet the requirement for investigating more complex social interactions of more than two mice.

\ifCLASSOPTIONcaptionsoff
  \newpage
\fi

\bibliographystyle{IEEEtran}
\bibliography{mybibfile}

\clearpage
\onecolumn

\setcounter{table}{0}
\setcounter{algorithm}{0}
\setcounter{figure}{0}
\setcounter{equation}{0}
\setcounter{page}{1}
\renewcommand\thefigure{S\arabic{figure}}
\renewcommand\thetable{S\arabic{table}}
\renewcommand\thealgorithm{S\arabic{algorithm}}

\section*{Supplementary A}

\textbf{Preliminaries.} ST-GCN \cite{yan2018spatial} is the first work adopting
Graph Convolutional Networks for skeleton data modelling. It is constructed by stacked spatio-temporal blocks, each of which is composed of a spatial convolution (GCN) block, followed by a temporal convolution (TCN) block. The spatial module utilizes the GCN to model the structural dependencies of nodes, which is formulated as:
\begin{equation}
\begin{split}
\mathbf{X}_{t}^{(l+1)}  = \sum_{k}^{K_{v}} \mathbf{W}_{k}\left ( \mathbf{X}_{t}^{l}  \mathbf{A}_{k}\right )
\label{equation:stgcn}
\end{split}
\end{equation}
where $K_{v}$ denotes the kernel size. $l$ is the layer index of the GCN. $\mathbf{W}_{k}$ is a trainable weight matrix that is implemented as $C_{\text {out }} \times C_{\text {in }} \times 1 \times 1$ convolution operation, where $C_{\text {out }}$ and $C_{\text {in}}$ are the output and input channels. $\mathbf{A}_{k}=\mathbf{\Lambda }_{k}^{-\frac{1}{2}}\tilde{\mathbf{A}}_{k}\mathbf{\Lambda }_{k}^{-\frac{1}{2}}$, where $\tilde{\mathbf{A}}_{k}$ is the adjacency matrix of the skeleton graph indicating intra-skeleton connections. $\mathbf{\Lambda }_{k}$ is the diagonal matrix, where $\mathbf \Lambda _{k}^{ii}=\sum_{j}\left(\tilde{\mathbf{A}}_{k}^{i j}\right)+c$, and $c$ is a small constant avoiding empty rows. On the temporal dimension, TCN is implemented by applying a $K_{t} \times 1$ 2D convolution operation to the input $\mathbf{X}  \in \mathbb{R}^{C \times T \times N}$ with $(T,N)$ dimensions, where $K_{t}$ is the kernel size.

The structure of the skeleton graph shown in Eq. (\ref{equation:stgcn}) is predefined by a fixed adjacency matrix. In order to learn an adaptive topology, \cite{shi2019two} presented the Adaptive Graph Convolutional Network (A-GCN), in which the adjacency matrix is divided into three complementary parts, as shown in Eq. (\ref{equation:2s-agcn}):
\begin{equation}
\begin{split}
\mathbf{X}_{t}^{(l+1)}  = \sum_{k}^{K_{v}} \mathbf{W}_{k} \mathbf{X}_{t}^{l}\left (\mathbf{A}_{k}+\mathbf{B}_{k}+\mathbf{C}_{k}\right )
\label{equation:2s-agcn}
\end{split}
\end{equation}
where $\mathbf{A}_{k}$ is the same as the one shown in Eq. (\ref{equation:stgcn}), which represents the physical structure of human body. $\mathbf{B}_{k}$ can be learned according to the training data and its elements can be an arbitrary value. It indicates the existence and strength of the connections between two nodes. $\mathbf{C}_{k}$ determines the connection strength between two nodes by calculating their similarity using the normalised embedded Gaussian function.

\begin{algorithm}[H]
\caption{Interaction-aware Transformer (IAT) reasoning process.}
\begin{algorithmic}[1]
\renewcommand{\algorithmicrequire}{\textbf{Input:}}
\renewcommand{\algorithmicensure}{\textbf{Output:}}
\REQUIRE Spatial-temporal node-level representation embedded with intra-, inter- and cross-skeleton interactions, i.e., $\mathbf X_{s_{i}}^{\text{cross},l}$ shown in Eq. (\ref{equation:after_cross-skeleton}).
\ENSURE Graph-level representation $\widetilde{\mathbf X}_{g}^{\text{cross}}$ for social behaviour classification.
\FOR {$l \leftarrow 1 $ to $ L $}
    \FOR {$i \leftarrow 1 $ to $ I $}
        \FOR {$j \leftarrow 1 $ to $ J-1 $}
        \STATE \small $\mathbf H_{s_{ij+1},m_{j+1}}^{\text{cross},l}  \leftarrow  LN(\mathbf Q_{s_{ij+1},m_{j+1}}^{l} + \mathbf \Upsilon _{m_{j+1}}(\mathbf X_{s_{ij},m_{j}}^{\text{cross},l}))$ (Eq. (\ref{equation: selfatt}));
        \STATE $\mathbf X_{s_{ij+1},m_{j+1}}^{\text{cross},l}  \leftarrow LN(\mathbf H_{s_{ij+1},m_{j+1}}^{\text{cross},l}+\mathbf \Gamma (\mathbf H_{s_{ij+1},m_{j+1}}^{\text{cross},l}) )$  (Eq. (\ref{equation: selfatt2}));
        \ENDFOR
        \STATE \small $SAP(\mathbf X_{s_{i1}}^{\text{cross},l})   \leftarrow \frac{1}{N_{s_{i}}^{1}} \sum_{m_{1} = 1}^{N_{s_{i}}^{1}}\mathbf X_{s_{i1},m_{1}}^{\text{cross},l} $ ;
        \STATE $ SMP(\mathbf X_{s_{i1}}^{\text{cross},l})  \leftarrow  max_{m \in \mathcal N_{s_{i}} ^{1}}(\mathbf X_{s_{i1},m_{1}}^{\text{cross},l}) $;
        \STATE $IAT(\mathbf X_{s_{i1}}^{\text{cross},l})  \leftarrow  \mathbf X_{s_{iJ},m_{J}}^{\text{cross},l}$ (w.r.t Eqs. (\ref{equation: selfatt}) and (\ref{equation: selfatt2}));
        \ENDFOR
         \STATE \small $\widetilde{\mathbf X}_{g}^{\text{cross},l} \leftarrow IAT([SAP(\mathbf X_{s_{1 1}}^{\text{cross},l}); \cdots ;IAT(\mathbf X_{s_{I1}}^{\text{cross},l})] )$ (Eq. (\ref{equation: multigraph2}));
         \IF{$l<L$}
         \STATE $\widetilde{\mathbf X}_{s_{iJ}}^{\text{cross},l} \leftarrow IAT(\widetilde{\mathbf X}_{g}^{\text{cross},l})$ (\ref{equation: multigraph_decoder}));
         \STATE $\mathbf X_{s_{i}}^{l}=\widetilde{\mathbf X}_{s_{iJ}}^{\text{cross},l}+\mathbf X_{s_{i}}^{\text{cross},l}$;
         \STATE $\mathbf X_{s_{i}}^{\text{cross},(l+1)}=CS-NLI(\mathbf X_{s_{i}}^{l})$ (w.r.t Eqs. (\ref{equation:gcn-tcn}), (\ref{equation:after_inter}) and (\ref{equation:after_cross-skeleton}));
         \ENDIF
         \ENDFOR
         \STATE $\widetilde{\mathbf X}_{g}^{\text{cross}} \leftarrow[\widetilde{\mathbf X}_{g}^{\text{cross},1};\widetilde{\mathbf X}_{g}^{\text{cross},2};\cdots ;\widetilde{\mathbf X}_{g}^{\text{cross},L}]$
\RETURN  Final graph-level representation $\widetilde{\mathbf X}_{g}^{\text{cross}}$.
\end{algorithmic}
\label{Alg:IAT}
\end{algorithm}

\textbf{More details about graph-level representation enhancement in IAT.} Given the representation of the first subgraph on the $l$-th layer of our network, i.e., $\mathbf X_{s_{i1}}^{\text{cross},l}$, we first calculate the average and maximum values of the representation in the spatial domain by spatial average pooling \cite{shi2019two} $SAP(\cdot)$  (see Algorithm \ref{Alg:IAT}) and max pooling \cite{zhang2020semantics} $SMP(\cdot)$, respectively. Eqs. (\ref{equation: selfatt}) and (\ref{equation: selfatt2}) can be treated as the implementation of function $IAT(\cdot)$ that describes the graph-level representation. Instead of fusing different graph-level representations across skeletons by direct summing over the spatial dimension, we attempt to model the relations between them using our proposed interaction-aware self-attention module to adaptively enhance the graph-level representation, formulated as follows:
\begin{equation}
\begin{split}
\begin{aligned}
\widetilde{\mathbf X}_{g}^{\text{cross},l}=&IAT(\mathbf Z_{s_{i1}}^{\text{cross},l}) \in \mathbb{R}^{ {C}_{l}\cdot{T}_{l}}
\\
\mathbf Z_{s_{i1}}^{\text{cross},l}=&[SAP(\mathbf X_{s_{1 1}}^{\text{cross},l});SMP(\mathbf X_{s_{11}}^{\text{cross},l});\\& \cdots ;IAT(\mathbf X_{s_{21}}^{\text{cross},l})] \in \mathbb{R}^{6 \times {C}_{l}\cdot{T}_{l}}
\end{aligned}
\label{equation: multigraph2}
\end{split}
\end{equation}
where $\widetilde{\mathbf X}_{g}^{\text{cross},l}$ is the enhanced graph-level representation that fuses various semantic information. $\mathbf Z_{s_{i1}}^{\text{cross},l}$ is the fused representation, including 3 types of graph-level representation at each skeleton branch.

\textbf{More details about decoder in IAT.} In most existing work \cite{yan2018spatial,shi2019two}, the node-level representation of one GCN-TCN block is directly fed into the next block for deeper spatio-temporal representation encoding, where the graph-level representation can be generated based on the last node-level representation. Different from these standard work, we add a decoder to the end of the encoder to adaptively update the node-level representation before sending the representation to the next layer. We  directly infer the node-level representation from the graph-level representation using our proposed interaction-aware self-attention presented in Section \ref{section3.3.1}:

\begin{equation}
\begin{split}
\begin{aligned}
\widetilde{\mathbf X}_{s_{iJ}}^{\text{cross},l}=&IAT(IAT(\mathbf Z_{s_{i1}}^{\text{cross},l}))\in \mathbb{R}^{N_{s_{i}} \times {C}_{l}\cdot {T}_{l}}
\end{aligned}
\label{equation: multigraph_decoder}
\end{split}
\end{equation}
where we define $J$ subgraphs for the decoder and the last one is $\widetilde{\mathbf X}_{s_{iJ}}^{\text{cross},l}$. Hence, the node-level representation for the $l$-th layer can be updated by $\mathbf X_{s_{i}}^{l}=\widetilde{\mathbf X}_{s_{iJ}}^{\text{cross},l}+\mathbf X_{s_{i}}^{\text{cross},l}$.

\textbf{Classification loss.} The classification loss is defined as:
\begin{equation}
\begin{split}
\begin{aligned}
\widetilde {\mathbf Y} &= Softmax(f_{o}(\widetilde{\mathbf X}_{g}^{\text{cross}}))
\\ \mathcal{L}_{\text{class}}&=-\frac{1}{B} \sum_{i=1}^{B} \sum_{j=1}^{C} \mathbf{Y}_{j}^{(i)} \log \widetilde{\mathbf{Y}}_{j}^{(i)}
\end{aligned}
\label{equation: outputgraphloss}
\end{split}
\end{equation}
where $f_{o}(\cdot)$ is a fully connected layer. $\widetilde{\mathbf X}_{g}^{\text{cross}}$ represents the final representation for classification, which is constructed by concatenating the graph-level representations of different layers.  $\widetilde{\mathbf{Y}}_{j}^{(i)}$ represents the predicted probability that the spatio-temporal skeleton graph with feature $\mathbf{X}^{(i)}$ belongs to class $j$, and $\mathbf{Y}_{j}^{(i)}$ is the corresponding ground truth. $B$ and $C$ denote the numbers of sliding windows and classes, respectively.

\clearpage
\onecolumn
\section*{Supplementary B}

\begin{figure*}[htbp]
\begin{center}
\includegraphics[width=8.4cm]{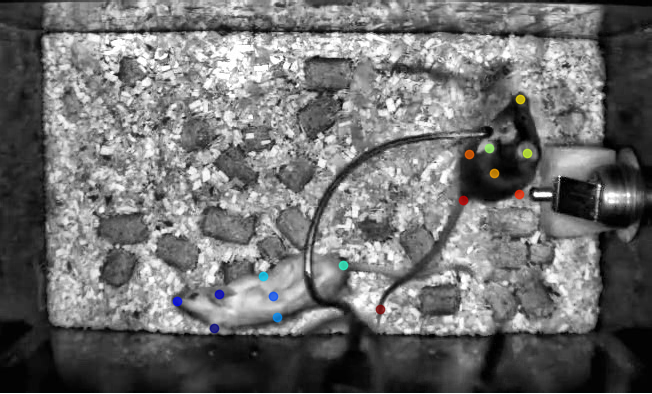}
\hspace{4mm}
\includegraphics[width=8.4cm]{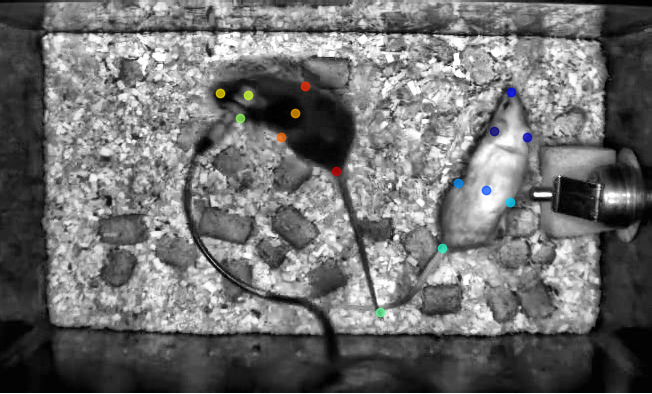}\\
% \hspace{-8mm}
(a) CRIM13-Skeleton\\
\includegraphics[width=8.4cm]{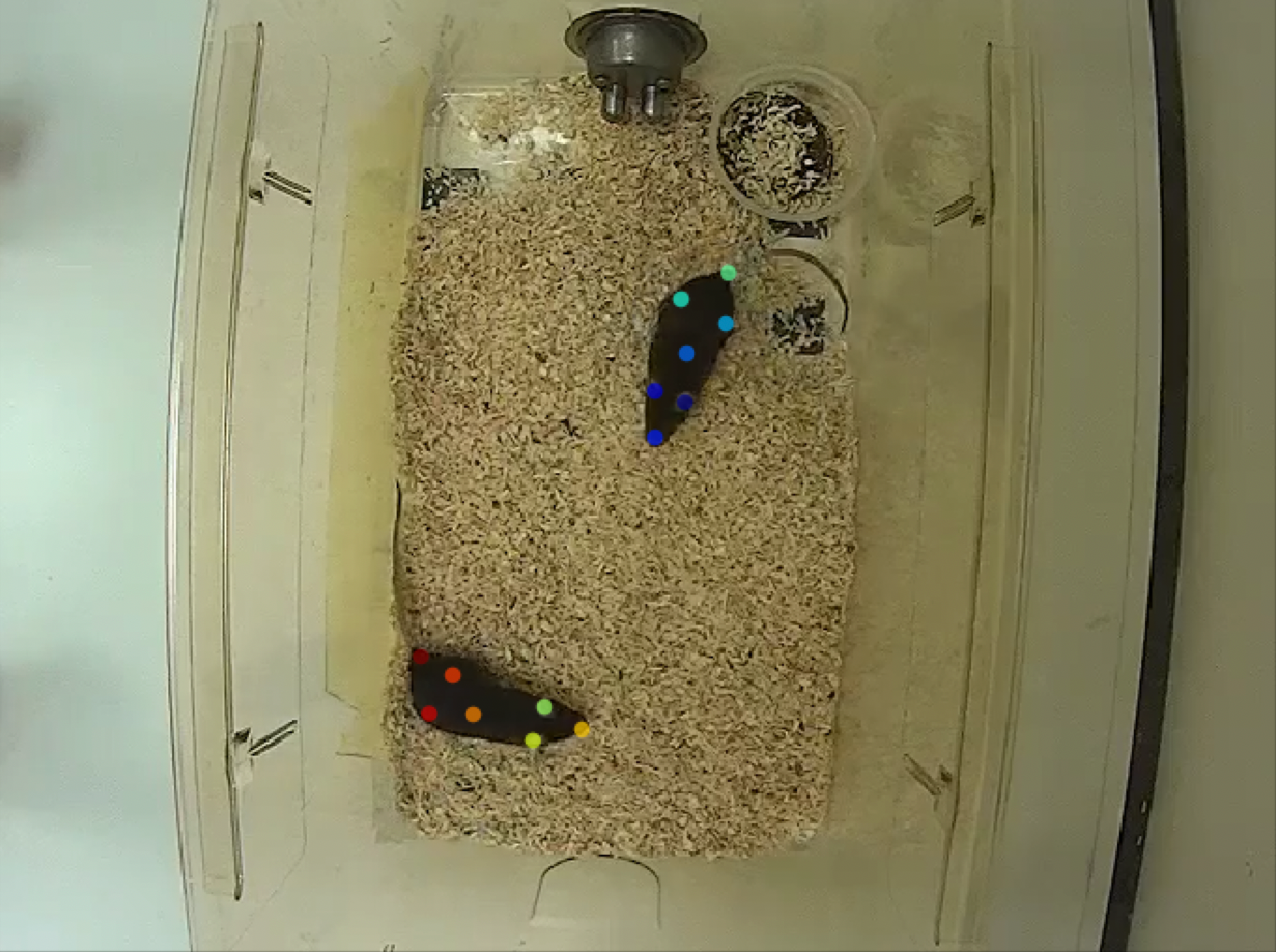}
\hspace{4mm}
\includegraphics[width=8.4cm]{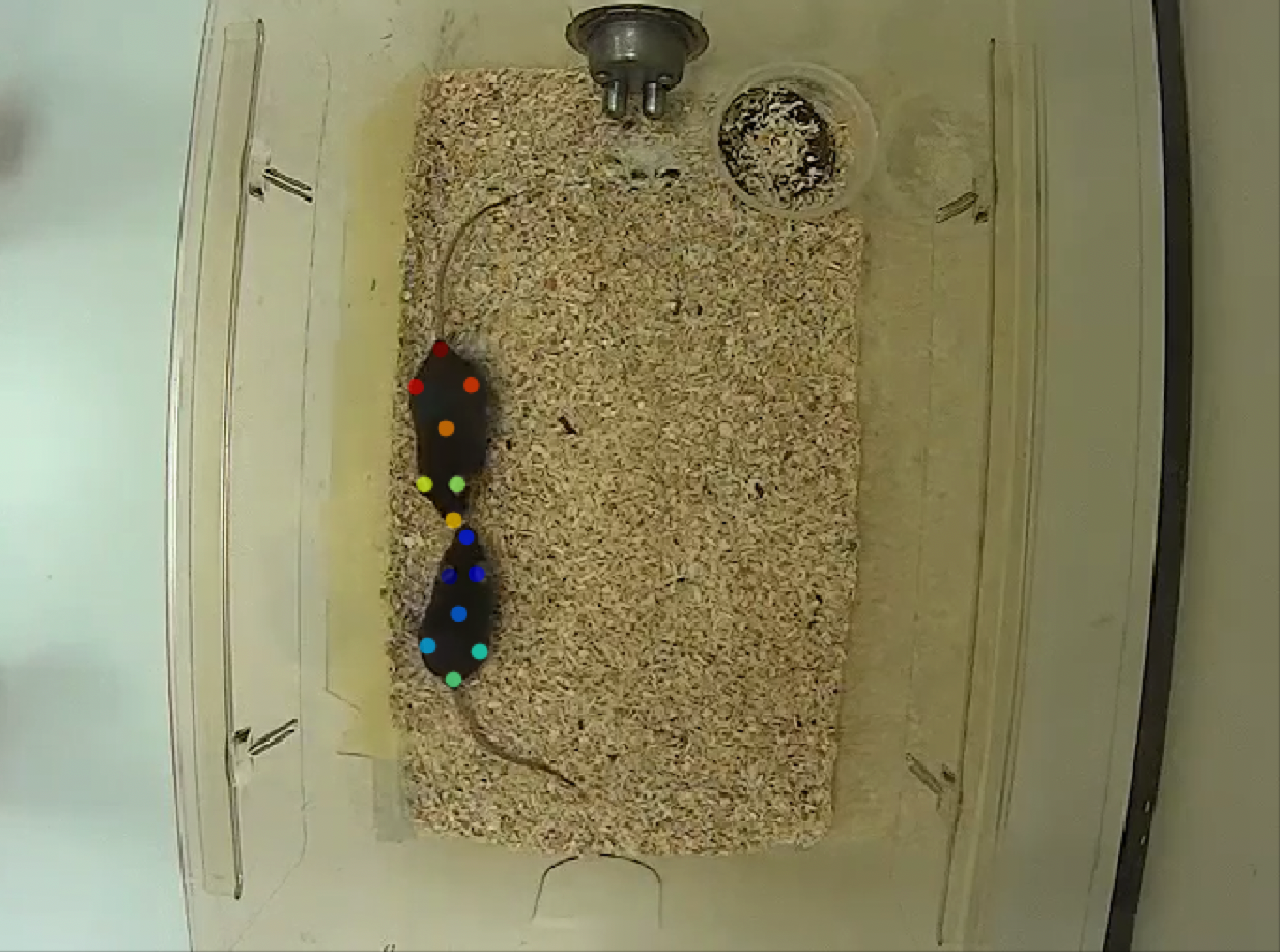}\\
% \hspace{-8mm}
(b) PDMB-Skeleton
\end{center}
\caption{Annotated locations of different mouse keypoints on the CRIM13-Skeleton and PDMB-Skeleton datasets. (a) The public CRIM13-Skeleton dataset \cite{nilsson2020simple} contains 8 keypoints for each mouse (i.e., 0-left ear, 1-right ear, 2-snout, 3-centroid, 4-left lateral, 5-right lateral, 6-tail base and 7-tail end). The numbers represent the order which the body-parts were annotated. (b) To establish our PDMB-Skeleton dataset, we extract frames every 500ms for each video, and all the extracted video frames were annotated using a freeware DeepLabCut (available at \href{https://github.com/DeepLabCut/DeepLabCut}{https://github.com/DeepLabCut/DeepLabCut}). A team of five professionals were trained to annotate the keypoints of each mouse. Similar to the CRIM13-Skeleton dataset, we annotate the locations of 7 body parts (i.e., 0-left ear, 1-right ear, 2-snout, 3-centroid, 4-left hip, 5-right hip and 6-tail base) for each mouse. We do not annotate the tail end because this keypoint is often occluded and the mouse tail is highly deformable in videos. We only use 7 body parts in all experiments on both PDMB-Skeleton and CRIM13-Skeleton datasets. In particularly, we ensure that the identity of each mouse remains unchanged during the process of annotation.  }
\label{fig:img_crim13}
\end{figure*}

\textbf{Dataset Construction.} We adopted a careful dataset construction process for our PDMB dataset (see Fig. S1). Specifically, we did not directly utilise a pre-trained model from DeepLabCut \cite{Mathis2018} to generate keypoint positions. Instead, we took the following steps: (1) Partial Frame Extraction: We initially extracted partial frames (at intervals of 500ms) from each video in the dataset. (2) DeepLabCut Annotation: The selected frames were then annotated using the DeepLabCut tool to manually label the positions of keypoints. (3) PDMB Training Set Construction: Subsequently, we utilised these annotated frames to construct the training set for PDMB. (4) Pose Estimation Network Training:  DeepLabCut was trained for mouse pose estimation on the PDMB dataset. (5) Keypoint Estimation: Finally, we used the pretrained pose estimation model to generate keypoint data (including confidence scores) for every frame in the dataset. 

Despite the presence of estimation errors, each keypoint is associated with a confidence score that quantifies this error. This confidence score is leveraged as a feature for each keypoint during network training, allowing the model to account for and learn from the uncertainties in the keypoint positions. This approach aligns with methodologies similar to SimBA \cite{nilsson2020simple}, which also utilises DeepLabCut for keypoint labelling. Additionally, in the case of the public dataset CRIM13-Skeleton, a confidence score is also included to measure position errors for each keypoint.

\textbf{Data Annotation.} Unlike the annotation method for CRIM13-Skeleton, we chose to annotate the left hip and right hip positions for the PDMB-Skeleton dataset. We referenced MARS \cite{segalin2021mouse} for mouse keypoint location annotation, labelling the left hip and right hip positions. The reason behind this choice is that the hip positions serve as crucial connectors between the upper body and the tail of the mouse. Considering the holistic perspective, the distribution of these seven keypoints, including the hip positions, is expected to provide a more comprehensive representation of the mouse's overall body structure. This can potentially contribute to a more nuanced understanding of mouse behaviour.

It's worth noting that the lack of a standardized mouse keypoint annotation scheme in the research community, including questions about which positions to annotate and the number of keypoints to include, poses a challenge. Different keypoint configurations may impact behaviour analysis. We acknowledge this limitation and plan to deal with this problem in our future work.

% \begin{figure*}[htbp]
% \begin{center}
% \begin{tabular}{ccc}
% \includegraphics[width=10.4cm]{img_crim13.png}&
% \hspace{-15mm}
% \includegraphics[width=10.4cm]{img_pdmb.png}\\
% % \hspace{-8mm}
% (a) CRIM13-Skeleton &(b) PDMB-Skeleton
% \end{tabular}
% \end{center}
% \caption{Illustrations of behaviours in CRIM13-Skeleton and our PDMB-Skeleton datasets: (a) CRIM13-Skeleton consists of 12 mouse behaviour classes. Each scenario records two freely behaving mice with 16 annotated keypoints from top-view. (b) Our PDMB-Skeleton dataset contains 9 mouse behaviours. Each scenario also records two freely behaving mice 14 annotated keypoints from top-view. In CRIM13-Skeleton, ‘drink’, ‘chase’ and ‘circle’ are the least represented behaviours with 0.4$\%$, 0.5$\%$ and 1.3$\%$, respectively. Similarly, ‘eat’, ‘circle’ and ‘chase’ together account for less than 2 of all the frames in our dataset. Such imbalance presents numerous challenges to automated recognition methods in both training and prediction. Thus, to alleviate the effect caused by the imbalance data, we use the PyTorch sampler, i.e., ImbalancedDatasetSampler (available at \href{https://github.com/ufoym/imbalanced-dataset-sampler}{https://github.com/ufoym/imbalanced-dataset-sampler}), to rebalance the class distributions when sampling from the imbalanced dataset and estimate the sampling weights automatically in all experiments. }
% \label{fig:behaviouroccurrence}
% \end{figure*}

\clearpage
\onecolumn
\section*{Supplementary C}

\begin{table*}[htbp]
\begin{center}
\caption{Ethogram of the observed behaviours, derived from CRIM13 \cite{burgos2012social}}
\label{tab:behaviour}
\begin{tabular}{p{2cm}|p{15cm}}
\toprule
Behaviour\quad & Description \\
\midrule
approach  & Moving toward another mouse in a straight line without obvious exploration.\\
attack & Biting/pulling fur of another mouse.\\
copulation & Copulation of male and female mice.\\
chase & A following mouse attempts to maintain a close distance to another mouse while the latter is moving.\\
circle &Circling around own axis or chasing tail\\
drink & Licking at the spout of the water bottle\\
eat & Gnawing/eating food pellets held by the fore-paws.\\
clean &Washing the muzzle with fore-paws (including licking fore-paws) or grooming the fur or hind-paws by means of licking or chewing.\\
human & Human intervenes with mice.\\
sniff & Sniff any body part of another mouse.\\
up & Exploring while standing in an upright posture.\\
walk away & Moving away from another mouse in a straight line without obvious exploration.\\
other & Behaviour other than defined in this ethogram, or when it is not visible what behaviour the mouse displays.\\

\bottomrule
\end{tabular}
\end{center}
\end{table*}

\begin{figure*}[htbp]
\begin{center}
\includegraphics[width=16.2cm]{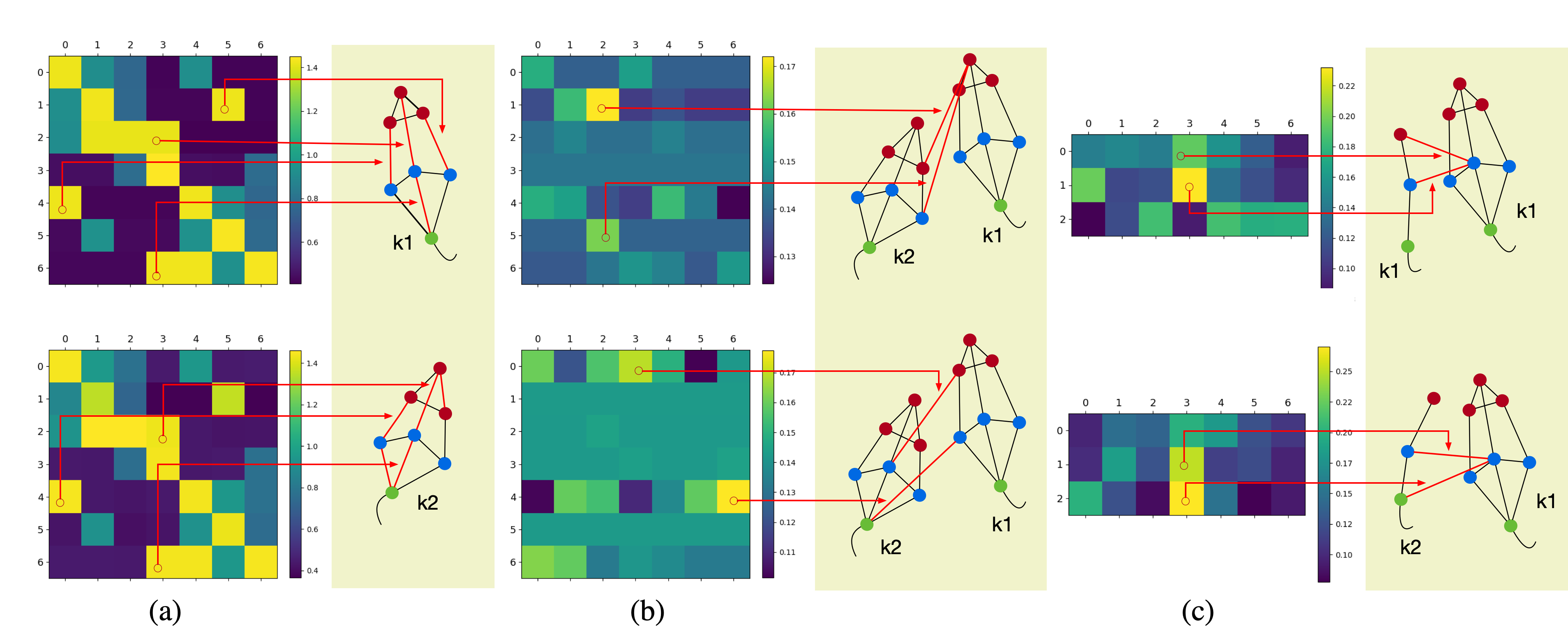}
\end{center}
\caption{Visualisation of the learned topologies of a social behaviour sample 'approach' on the CRIM13-Skeleton dataset (at the beginning of training, i.e., epoch=10). (a) The topologies representing intra-skeleton interactions of mouse $k_{1}$ (top) and $k_{2}$ (bottom). The number of keypoints V is 7 and its configuration is shown in Fig. \ref{fig:img_crim13}. Here, we show the summation of the learned topologies on the three subsets generated by the partition strategy \cite{yan2018spatial}. (b) The topologies of bidirectional inter-skeleton interactions learned by our model, i.e., $\mathbf{A}_{k_{1}\to k_{2}}^{l=1}$ (top)  and $\mathbf{A}_{k_{2}\to k_{1}}^{l=1}$ (bottom). (c) The topologies of cross-skeleton interactions ($s_{1}$ to $s_{2}$) learned by our model, i.e., $\mathbf{A}_{s_{1}\to s_{2}}^{l=1}$ (top)  and $\widetilde{\mathbf{A}}_{s_{1}\to s_{2}}^{l=1}$ (bottom). For each type, we use red lines to indicate the interactions with high significance. We observe that the module generates relatively dense fully connected graph at the beginning of training, especially for the inter- and -cross interactions, i.e., interactions not related to behaviours.
On the contrary, our final module (Fig. \ref{fig:matrix}) tends to give less attention to trivial interactions. }
\label{fig:matrix_inter}
\end{figure*}

\begin{figure*}[htbp]
\begin{center}
\begin{tabular}{ccc}
\includegraphics[width=9.6cm]{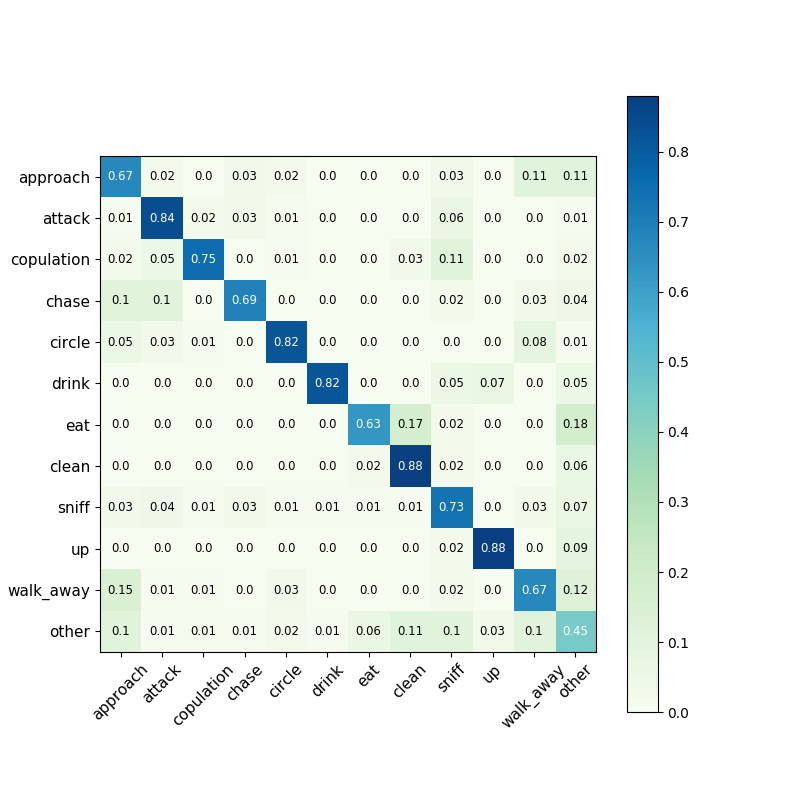}&
\hspace{-15mm}
\includegraphics[width=9.6cm]{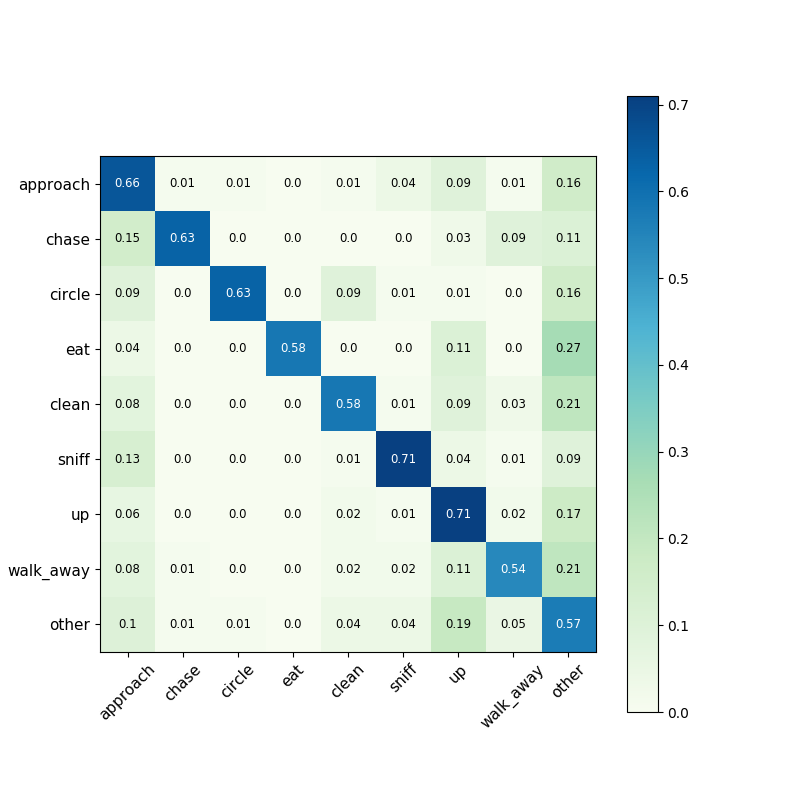}\\
% \hspace{-8mm}
(a) CRIM13-Skeleton &(b) PDMB-Skeleton
\end{tabular}
\end{center}
\caption{Confusion matrices of our method (i.e., CS-IGANet) on CRIM13-Skeleton (a) and PDMB-Skeleton datasets (b). The diagonal cells in each confusion matrix show the percentage of correct classifications. The confusion matrix is obtained for measuring the agreement between the ground-truth (row) and the predicted labels (column). The non-diagonal cells contain the percentages of the incorrectly classified behaviors. In each row, all the values should sum to be 1. The higher probabilities along the diagonal and the lower off-diagonal values indicate the degrees of successful classification for all the categories. The colour bar indicates the degree of the agreement whilst deep blue indicates the agreements close to 100$\%$.}
\label{fig:confusionm}
\end{figure*}

\begin{table*}
\begin{center}
\caption{
Ablation experiments for the Cross-Skeleton Node-level Interaction (CS-NLI) module on the CRIM13-Skeleton dataset. We present the classification accuracy ($\%$) of each behaviour,  average accuracy over all the behaviours, FLOPs and parameter number. The best performance is highlighted in bold.}
\label{tab:CS-NLI_two_densegraphs}
\begin{tabular}{p{3.2cm}p{0.8cm}p{0.4cm}p{1cm}p{0.4cm}p{0.4cm}p{0.4cm}p{0.4cm}p{0.4cm}p{0.4cm}p{0.4cm}p{0.4cm}p{0.4cm}p{0.7cm}p{0.7cm}p{0.7cm}}
\toprule
Methods\quad & Approach & Attack & Copulation & Chase & Circle & Drink & Eat & Clean &Sniff & Up & Walk away & Other &Average & Params &FLOPs\\
\midrule
CS-NLI(two dense graphs) & 63.35	&77.91	&72.02	&40.91	&59.79	&\textbf{79.93}	&\textbf{62.35}	 &83.52	&62.73	&84.65	&55.75	&\textbf{50.67} & 66.13 &3.41M  & 0.40G\\
CS-NLI(multi-scale graphs) & \textbf{69.24}	&\textbf{81.81}	&\textbf{78.91}	&\textbf{44.73}	&\textbf{66.23}	&79.12	&57.70	&\textbf{86.87}	&\textbf{65.72}	&\textbf{85.12}	&\textbf{59.10}	&45.02 &\textbf{68.30} & 2.90M  & 0.37G\\
\bottomrule
\end{tabular}
\end{center}
\end{table*}

\hspace{-3cm}

\begin{table}
\centering
\begin{threeparttable}

\caption{Comparisons with state-of-the-art methods on the NTU-Interaction and NTU120-Interaction datasets in accuracy ($\%$).}
\label{tab:comparison_human}
\begin{tabular}{c|cc|cc}
\hline
 \multirow{2}{*}{Method} & \multicolumn{2}{c|}{NTU-Interaction} &  \multicolumn{2}{c}{NTU120-Interaction} 
\\ \cline{2-5} 
& X-Sub & X-View & X-Sub & X-View\\ 
\hline
ST-GCN* \cite{yan2018spatial}& 89.31 & 93.72 & 80.69 & 80.27\\
2S-AGCN* \cite{shi2019two}& 93.36 & 96.67 & 87.83 & 89.21\\
CTR-GCN* \cite{chen2021channel}& 95.31 & 97.60 & 92.03 & 92.82\\
2S-DRAGCN* \cite{zhu2021dyadic}& 94.68 & 97.19 & 90.56 & 90.43\\
2P-GCN* \cite{li2022two}& 97.05 &\textbf{98.80} & 93.47 & 93.73\\
Ours\textsuperscript{+} (CS-IGANet) &  \textbf{97.12} &  97.89& \textbf{94.39} & \textbf{94.70}\\
\hline
\end{tabular}
\begin{tablenotes} 
\item[*] Results are reported in \cite{li2022two}. 
\item[+] We follow the same pre-processing method as described in \cite{li2022two}. We adopt multi-scale skeleton graphs composed of 25 joints (for each actor) and 12 joints \cite{qi2023semantic}, respectively, to serve as dense and sparse skeleton graphs in our framework.
\end{tablenotes} 
\end{threeparttable}
\end{table}

\begin{figure*}
\begin{center}
\includegraphics[width=16.2cm]{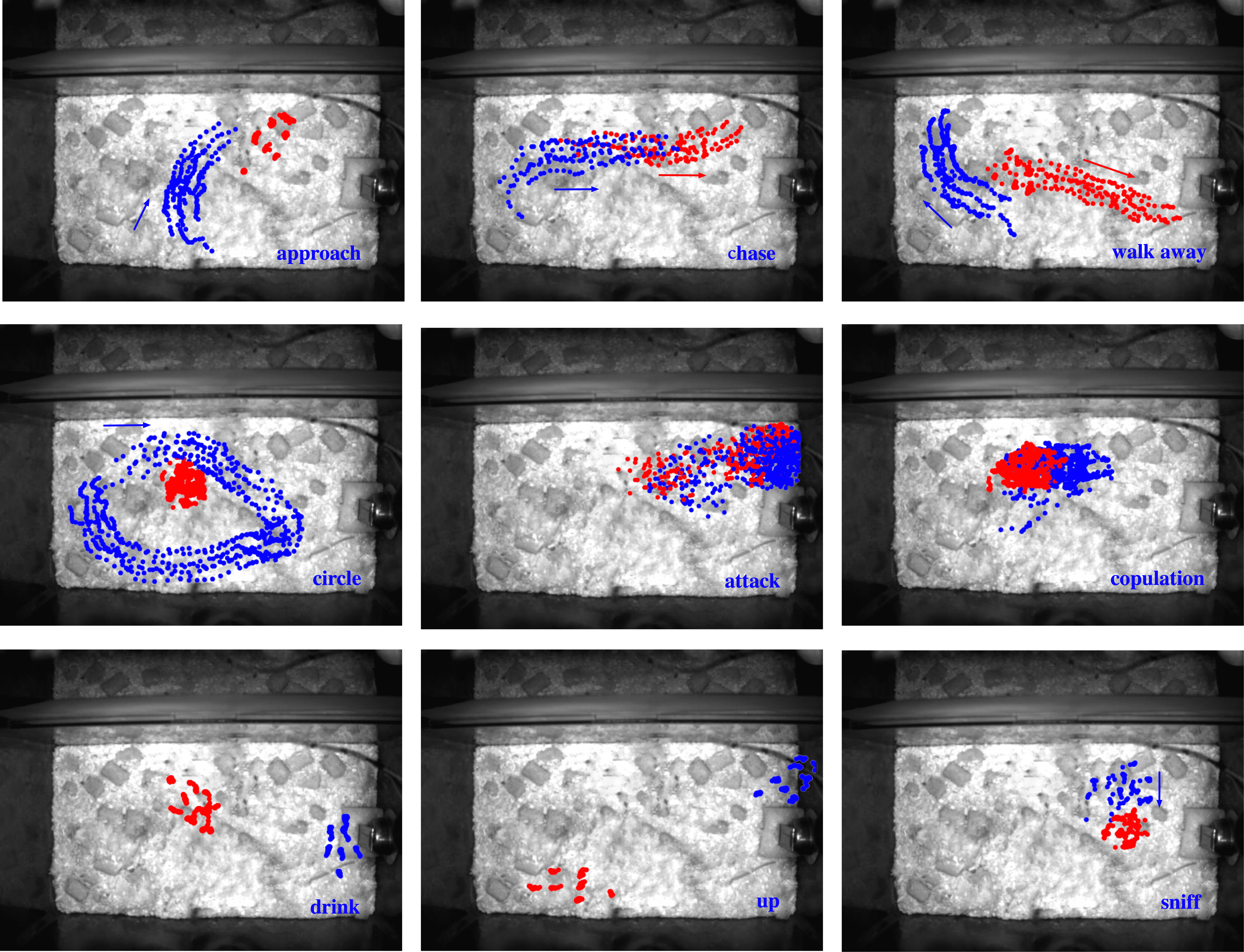}
\end{center}
\caption{Examples of motion trajectory of different behaviours in the CRIM13-Skeleton dataset. Blue and red points indicate the keypoints of the resident mouse and the intruder, respectively. Blue and red arrows represent the direction of motion. For some behaviours (e.g., clean and up) without significant movement, we do not give the direction of motion.  }
\label{fig:trajectory_eachbehaviour}
\end{figure*}

% \begin{figure*}
% \begin{center}
% \includegraphics[width=16.2cm]{trajectory_eachbehaviour2.png}
% \end{center}
% \caption{Examples of motion trajectory of different behaviours in the CRIM13-Skeleton dataset. Blue and red points indicate the keypoints of the resident mouse and the intruder, respectively. Blue and red arrows represent the direction of motion. For some behaviours (e.g., clean and up) without significant movement, we do not give the direction of motion.  }
% \label{fig:trajectory_eachbehaviour2}
% \end{figure*}

\clearpage
\onecolumn
% \end{flushleft}

% \section*{Supplementary D}

\end{document}